\documentclass{article}

\usepackage{microtype}
\usepackage{graphicx}
\usepackage{subfigure}
\usepackage{booktabs} % for professional tables
\usepackage{comment}

\usepackage{subcaption}

\usepackage{hyperref}
\usepackage{float}

\usepackage[accepted]{icml2024}

\usepackage{amsmath}
\usepackage{amssymb}
\usepackage{mathtools}
\usepackage{amsthm}
\usepackage{multirow}
\usepackage[capitalize,noabbrev]{cleveref}

\theoremstyle{plain}

\newtheorem{theorem}{Theorem}%[section]
\newtheorem{proposition}[theorem]{Proposition}

\theoremstyle{definition}

\theoremstyle{remark}
\newtheorem{remark}{Remark}

\usepackage[textsize=tiny]{todonotes}

\begin{document}

\twocolumn[
\icmltitle{TimeMIL: Advancing Multivariate Time Series Classification via a Time-aware Multiple Instance Learning}



\icmlsetsymbol{equal}{*}

\begin{icmlauthorlist}
\icmlauthor{Xiwen Chen}{equal,clemson}
\icmlauthor{Peijie Qiu}{equal,wasu}
\icmlauthor{Wenhui Zhu}{equal,asu}
\icmlauthor{Huayu Li}{ua}
\icmlauthor{Hao Wang}{clemson}\\
\icmlauthor{Aristeidis Sotiras}{wasu}
\icmlauthor{Yalin Wang}{asu}
\icmlauthor{Abolfazl Razi}{clemson}
\end{icmlauthorlist}

% \icmlaffiliation{clemson}{School of Computing, Clemson University, USA}
% \icmlaffiliation{wasu}{McKeley School of Engineering, Washington University in St. Louis, USA}
% \icmlaffiliation{asu}{School of Computing and Augmented Intelligence, Arizona State University, USA}
% \icmlaffiliation{ua}{Department of Electrical \& Computer Engineering, University of Arizona, USA }
\icmlaffiliation{clemson}{Clemson University, USA.}
\icmlaffiliation{wasu}{Washington University in St. Louis, USA.}
\icmlaffiliation{asu}{Arizona State University, USA.}
\icmlaffiliation{ua}{University of Arizona, USA }

\icmlcorrespondingauthor{Xiwen Chen}{xiwenc@g.clemson.edu}
\icmlcorrespondingauthor{Abolfazl Razi}{arazi@clemson.edu}

\icmlkeywords{Multivariate Time Series Classification, Weakly-supervised Learning}

\vskip 0.3in
]



 \printAffiliationsAndNotice{\icmlEqualContribution} % otherwise use the standard text.

% This document provides a basic paper template and submission guidelines.
% Abstracts must be a single paragraph, ideally between 4--6 sentences long.
% Gross violations will trigger corrections at the camera-ready phase.

\begin{abstract}
% Deep learning-based multivariate time series classification (MTSC) has recently achieved remarkable performance enhancements by incorporating the transformer, convolutional neural network, and other methodologies. However, these methods often employ a supervised learning scheme to fit classification tasks, overlooking the inherent sparsity of interest patterns within time series data. In this paper, we defined  MTSC as a weakly supervised learning problem. This not only aids the network in learning to provide interpretability but also models interdependencies within time series data to enhance performance. Based on above, we introduce multiple-instance learning (MIL) into the MTSC. However, due to the distinctive nature of time series data, it typically possesses properties characterized by temporal correlation and ordering interdependencies.. This is constrained by the invariant permutation defined in MIL. Here, we proposed a time data-specific- MIL method, TimeMIL. First, we formally discussed and proved the possibility of modeling the temporal correlation in MIL from the information theory perspective, offering a corresponding solution. Second, we also proposed a learnable wavelet positional encoding in TimeMIL, which aims to preserve the multi-scale time-frequency ordering information. As experimental results, our proposed method achieved state-of-the-art performance on 28 benchmark datasets compared with 26 latest baseline methods. Comprehensive analysis and visualization shed light on the effectiveness and potential of MIL in MTSC.

Deep neural networks, including transformers and convolutional neural networks, have significantly improved multivariate time series classification (MTSC). However, these methods often rely on supervised learning, which does not fully account for the sparsity and locality of patterns in time series data (e.g., diseases-related anomalous points in ECG). To address this challenge, we formally reformulate MTSC as a weakly supervised problem, introducing a novel multiple-instance learning (MIL) framework for better localization of patterns of interest and modeling time dependencies within time series. Our novel approach, TimeMIL, formulates the temporal correlation and ordering within a time-aware MIL pooling, leveraging a tokenized transformer with a specialized learnable wavelet positional token. The proposed method surpassed 26 recent state-of-the-art methods, underscoring the effectiveness of the weakly supervised TimeMIL in MTSC. The code will be available at \url{https://github.com/xiwenc1/TimeMIL}.

% \fancyfoot[5]{The camera-ready version will be available soon after further polishing.}

% \fbox{4-6 sentence, single paragraph}
% This paper proposes a time-aware multiple instance learning framework, termed \textit{TimeMIL}, designed for multivariate time series classification (MTSC). Different from the recent deep learning (DL)-based MTSC methods designed from the perspective of supervised learning and focusing on architecture optimization, TimeMIL, a pioneering framework, formulates multiple instance learning (MIL) problems for MTSC and addresses the limitations of standard multiple instance learning by effectively modeling temporal correlations and enhances interpretability. Consequently, we propose a transformer-like MIL aggregator equipped with a novel learnable Wavelet positional encoding to capture the temporal ordering of time data. Our intensive experiments demonstrate that TimeMIL significantly outperforms state-of-the-art methods across various datasets, including 26 comparison methods on 28 UEA datasets. This work not only advances the field of time series classification but also offers a new perspective on handling temporal data in weakly supervised learning settings.
\end{abstract}
% \vspace{-0.1in}
\section{Introduction}\label{sec:intro}
Time series data mining has witnessed considerable growth %as an area of research 
 in the last decade with numerous applications in  classification~\cite{tsclassification}, forecasting~\cite{timeforecasting}, and anomaly detection~\cite{tsabnormaldetection}. Particularly,
 multivariate time series classification (MTSC), which aims to assign labels to time sequences,  
 % time series classification (TSC), which aims to assign labels to time sequences, is crucial in scenarios involving time-dependent data, such as financial forecasting and Predictive Maintenance. % one of the most important tasks. 
 % In this work, we focus on multivariate time series classification (MTSC), which 
 is challenging but crucial in most real scenarios, such as healthcare~\cite{magnetoencephalography,tang2023detecting}, human action recognition~\cite{shokoohi2017generalizing,amaral2022summertime}, audio signal processing \cite{ruiz2021great}, Internet of Things~\cite{bakirtzis2022deep}, and semantic communication~\cite{zhao2023classification}.
 %\arr{add one ref for human action recognition and remove one from audio to be consistent}

%bagnall2018uea,
\begin{figure}[t]
    \centering
    \includegraphics[width=0.45\textwidth]{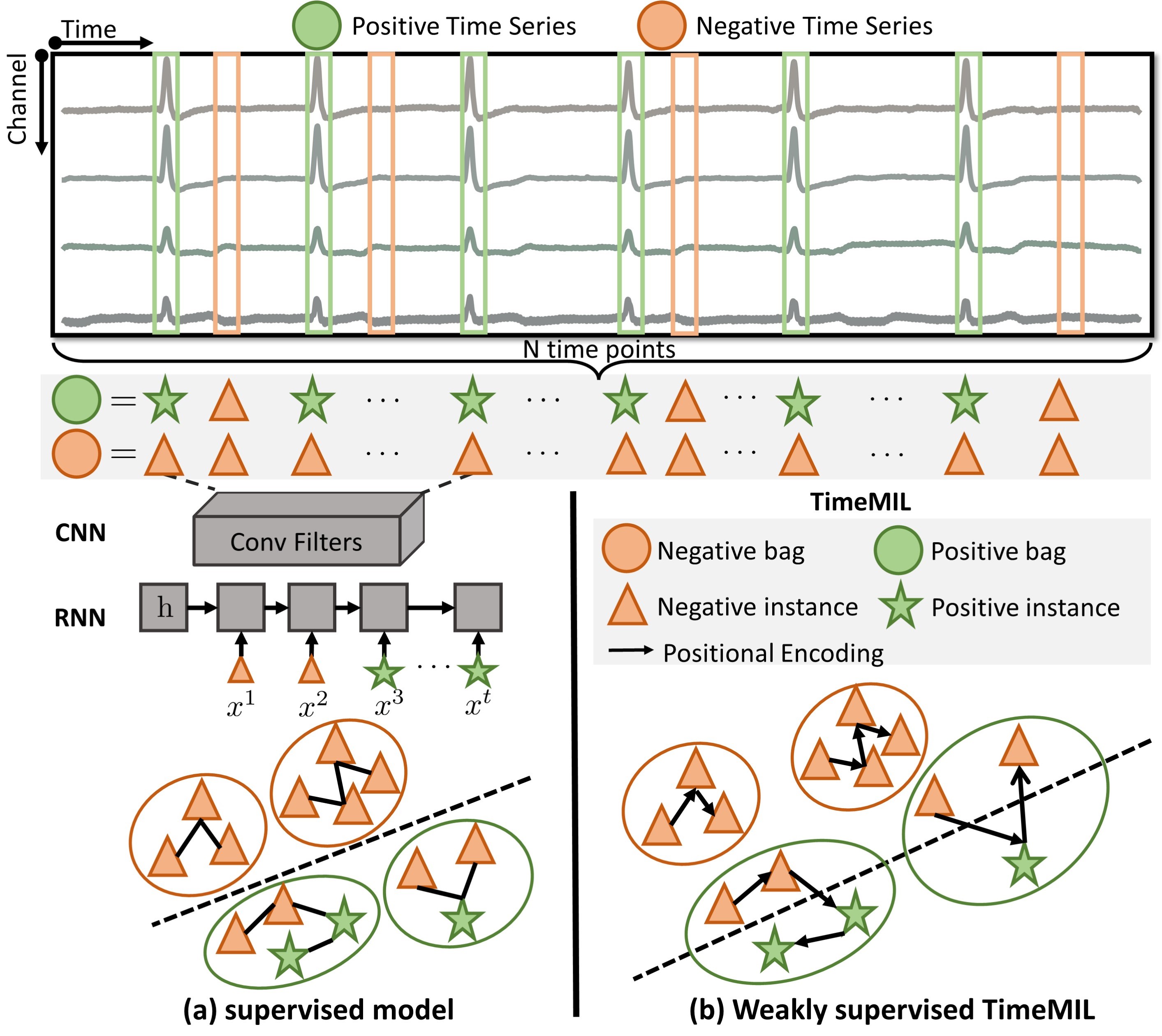}
    \vspace{-0.2cm}
    \caption{\textbf{(a):} The decision boundary of fully supervised methods is determined by assigning a label to each time series. \textbf{(b):} TimeMIL makes decisions by discriminating positive and negative instances in time series, where each time point is an instance, and its label is typically not available in reality.}
    % \arr{This figure is good. A few minor comments: 1-Define what is the data in the caption, like say " for EEG signal" after weekly supervised TimeMIL. 2-for ECG signal, we typically represent one full cycle as a sample or data point and then label it (like for arrhythmia classes), I am not sure about EEG but I guess there might be a similar concept there too. you are showing peaks with green and flat parts with orange, but each sounds like a portion of a cycle, not a full cycle. Maybe you can show a longer signal where both greens and oranges representing one full cycle, green for healthy and orange for issues....}}  
    \vspace{-0.1cm}
    \label{fig:concept}
\end{figure}

Recently, deep neural networks have achieved state-of-the-art performance in various time series tasks compared to traditional methods~\cite{seto2015multivariate,schafer2017multivariate,li2023sleep,tang2022omniscale,anonymous2023moderntcn,li2024continual}.  
Their popularity and success in time series modeling can be attributed to their automatic feature extraction in conjunction with inductive biases. To this end, the time series task is typically formulated as a fully supervised learning task by employing a wide variety of architectures, such as
% is effective in capturing local dependencies and patterns in a time sequence
recurrent neural networks (RNN)~\cite{rnnMTS1,rnnMTS2}, long short-term memory (LSTM),~\cite{karim2019multivariate,tang2022omniscale,karim2019multivariate}, convolution neural networks (CNN) ~\cite{zhang2020tapnet,ismail2020inceptiontime,wu2022timesnet,tang2022omniscale,anonymous2023moderntcn}, and transformers~\cite{zhou2022fedformer,zhang2023crossformer,nie2023a,li2024mts,li2024multi}.
Although these methods vary in their architectural biases for modeling time series, they share a common underlying strategy: dividing a time series into a set of time points and then modeling the local and global dependencies among them, such as changes over time and the presence of multiple periodic patterns. Importantly, patterns of interest in time series are typically sparse and localized~\cite{sparseandlocalize,sparseandlocalize2,sparseandlocalize3,sparseandlocalize4}, while the most discriminative time points within a time series are typically unknown due to their laborious annotation. This poses a significant challenge for fully supervised learning in accurately determining the decision boundary (see Fig.~\ref{fig:concept}(a)). Instead, given the inherent properties of time series, we formulate the MTSC tasks as a weakly supervised learning paradigm (see Fig.~\ref{fig:concept}(b)). 

% Compared with traditionally handcrafted feature extraction methods~\cite{seto2015multivariate,schafer2017multivariate}, Deep learning has dominated the time series based various tasks, which introduced the Convolution Neural Network (CNN)~\cite{zhang2020tapnet,ismail2020inceptiontime,wu2022timesnet,tang2022omniscale,anonymous2023moderntcn}, Long Short-Term Memory (LSTM)~\cite{karim2019multivariate,tang2022omniscale,karim2019multivariate}, Recurrent Neural Networks (RNN)~\cite{rnnMTS1,rnnMTS2} and transformer~\cite{zhou2022fedformer,zhang2023crossformer,nie2023a}. 
% % ~\cite{transformerMTS1,transformerMTS2,transformerMTS3}. 

Multiple instance learning (MIL) is a weakly supervised learning method that assigns a label to a collection of instances, known as a bag. This makes MIL a natural choice for the MTSC task by collectively treating each time point as an instance and an entire time series as a bag. Early attempts of MIL in time series relied on hand-crafted features and classic MIL models~\cite{stikic2011weakly,milmtsc}. In contrast, modern MILs which use deep neural networks to extract the feature automatically and consistently exhibit superior performance compared to the classic MILs with handcrafted features~\cite{wang2018revisiting,ilse2018attention,anonymous2024inherently}. However, standard MILs may fail %it also fails 
to capture correlations between instances, since standard MILs assume the independence and identical distribution of instances with a permutation-invariant property~\cite{ilse2018attention}.
% \arr{this sentence is a little ambiguous, you can use standard MILs or modern MILs or MILs in general instead of "they" to be accurate.}
In contrast, MTSC data typically exhibits temporal correlations and ordering dependencies, posing %which poses 
significant challenges for directly translating MIL into MTSC.

This paper introduces a generic MIL framework for time series, termed \textit{TimeMIL}. We address several limitations when using standard MIL methods, such as their failure to model the permutation information and temporal correlation among instances. We explore their necessity from an information-theoretic perspective, which suggests that modeling the permutation information and temporal correlation can lower the uncertainty of classification systems. To this end, we propose a time-aware MIL pooling, leveraging the self-attention mechanism and a novel learnable \textit{wavelet positional encoding}, where the former is used to capture the temporal correlation between instances, and the latter is used to characterize time ordering information. 

\noindent\textbf{Contributions:} \textbf{(i)} To the best of our knowledge, we are the first to formally formulate a generic MIL framework for multivariate and multi-class time series classification from an information-theoretic perspective. %view
\textbf{(ii)} We propose a Time-aware MIL pooling based on a tokenized transformer and a novel learnable \textit{wavelet positional encoding} (WPE) to model complex patterns within time series. %\textbf{(iii)}  %\arr{I would remove "(iii)" [I mean remove just the number, the sentence is good!] because the last one is not a contribution rather emphasizes our method is good. 2-add a word between  state-of-the-art and methods to be more specific.}
The proposed method outperforms 26 recent state-of-the-art methods in 28 datasets and offers inherent interpretability.

\section{Related Works}
% \fbox{fully revised}
% \vspace{-0.1in}
% \subsection{Multivariate Time Series Classification}
% The early DL approaches designed for MTSC often take benefit of LSTM and CNN
% \vspace{-0.1in}
\noindent\textbf{Multivariate Time Series Classification.} 
The recent DL methods specifically designed for MTSC can roughly be divided into two categories: (i) CNN/LSTM Hybrid architecture~\cite{karim2019multivariate,zhang2020tapnet}, where LSTM is often used to capture sequential dependencies and CNN is used to capture the local features. (ii) Purely CNN architecture~\cite{ismail2020inceptiontime,li2021shapenet,tang2022omniscale}, where long-term dependencies, short-term dependencies, and cross-channel dependencies are claimed to be captured by multiple kernels with varying kernel sizes. 

Recently, \textit{General Time Series Analysis Framework}~\cite{wu2022timesnet,anonymous2023moderntcn} has been also proposed for multiple mainstream tasks, including classification, imputation, short-term forecasting, long-term forecasting, and anomaly detection, with simple modifications.
 % . In fact, they also demonstrate that the recent models originally designed for forecasting have competitive performance for classification tasks with simple modifications.
% Similar to the Computer Vision (CV) and Natural language Processing (NLP) communities, 
Transformer-based models \cite{zhou2022fedformer,zhang2023crossformer,nie2023a} and MLP-based models \cite{zeng2023transformers,zhang2022less,li2023mts,li2023revisiting} have also been developed and improved over the last few years for this purpose due to their excellent scaling behaviors. Nonetheless, they did not yet fully replace CNN-based models, which continue to exhibit impressive performance \cite{liu2022scinet,wang2023micn,anonymous2023moderntcn}.

% Although the aforementioned methods have witnessed extensive use in time series analysis applications, \arr{say what is their weakness or what cases they are not useful for before saying our method is different} they differ fundamentally from our approach. These methods are rooted in supervised learning, focusing on optimizing their architectural designs (e.g., CNN and transformer). In contrast, our work introduces a weakly supervised TimeMIL, which provides a novel perspective to describe time series. \arr{this sentence above is not strong because being different by itself is not an advantage. You have to say why our new perspective/approach is better}
% In addition, the proposed TimeMIL is inherently interpretable.

% % \xc{@Dr. Razi, we mention their weakness in the introduction.}

Although the aforementioned methods have witnessed extensive use in time series analysis applications, as discussed in the introduction, these methods are rooted in supervised learning and often focus on optimizing their architectural designs (e.g., CNN and transformer), which still cannot solve the essential issue that they are challenging to determine the accurate decision boundary. In contrast, our work introduces TimeMIL, which provides a novel perspective to describe the decision boundary of time series in a weakly supervised view. In addition, the proposed TimeMIL is inherently interpretable.

\begin{figure*}[t]
    \centering
    \includegraphics[width=0.81\textwidth]{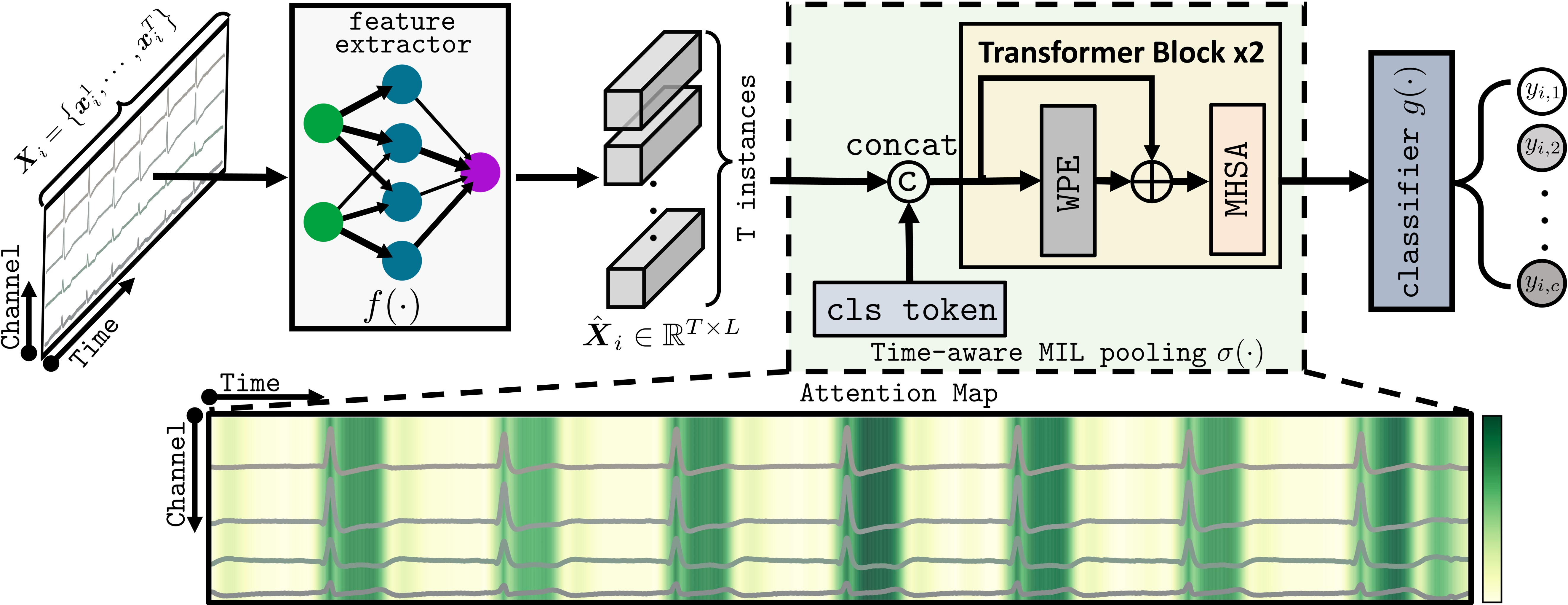}
    \vspace{-0.2cm}
    \caption{The proposed framework of TimeMIL for time series classification with enhanced interpretability: (i) a feature extractor to obtain instance-level feature embeddings, (ii) a MIL pooling to aggregate instance embeddings to a bag-level feature, embedding, and (iii) a bag-level classifier to map bag-level feature to a label prediction. Each time point is treated as an instance and the time series as a bag. Time ordering information and instance correlation are captured by taking the mutual benefit of WPE and MHSA in our TimeMIL pooling (highlighted in green).}  % The interpretability is offered by our TimeMIL pooling via a class token-related attention map.
    \vspace{-0.3cm}
    \label{fig:network}
\end{figure*}

 % \noindent\textbf{General Time Series Analysis Framework.} 

% \vspace{-0.1in}

% \fbox{Move the general definition of MIL here or to the Intro}. 

% MIL operates on a higher level of abstraction by dealing with sets of instances, collectively referred to as bags. Each bag is assigned a label that reflects the aggregate characteristics of its constituent instances, while the instances themselves remain unlabeled. 

% MIL is a weakly-supervised learning method~\cite{MILproblem}, where training data is organized into bags, each bag containing multiple instances. 
% The instances within a bag remain unlabeled, while each bag is assigned a label that reflects the aggregate characteristics of its instances.
\noindent\textbf{Multiple Instance Learning.} 
% \vspace{-0.08in}
MIL, a weakly supervised method, is widely used for histological image classification due to its advantage in localizing tumors within gigapixel images~\cite{ilse2018attention,li2021dual,zhang2022dtfd}. 
% The applications of Multiple Instance Learning (MIL) are divided into two main catagories: i) instance-based MIL and ii) embedding-based MIL. Instance-based methods involve assigning the bag-level label to each instance within the bag and then aggregating these instance-level predictions to form the final bag-level prediction. However, this approach has been found to be less effective due to the noise in instance-level supervision~\cite{wang2018revisiting}. In contrast, embedding-based methods first convert instances into feature embeddings and then aggregate these embeddings to determine the bag-level prediction, which has shown to be more effective. As discussed in Section \ref{}, MIL has been employed in several applications, such as ; 
However, the application of MIL to time series data has rarely been explored. %\cite{stikic2011weakly,milmtsc,zhu2021uncertainty,anonymous2024inherently}
An early exploration by~\cite{stikic2011weakly} applied classic multi-instance SVM~\cite{andrews2002support} to wearable sensor data. However, this method fails to model sequential dependencies among time points (instances). To address this limitation, \cite{milmtsc} proposes incorporating an autoregressive hidden Markov model into the MIL framework to model the dependencies between instances. However, both methods rely on FFT and hand-crafted statistical features. Alternatively, modern MILs typically employ deep neural networks for automatic feature extraction and aggregation~\cite{wang2018revisiting}, achieving superior performance over conventional instance-based MIL, like the one used in \cite{zhu2021uncertainty} for MTSC. 
% Authors in \cite{zhu2021uncertainty} apply instance-based MIL for TSC problem, which first predicts the class for each instance and then predicts the bag label by aggregating instance-level prediction. However, its performance is challenged by the noisy instance-level supervision~\cite{wang2018revisiting}.
Most notably, the attention-based MIL (ABMIL) is proposed by~\cite{ilse2018attention}, which makes MIL inherently interpretable by weighting each instance according to its importance. Since its invention, %inception, 
ABMIL has emerged as the standard paradigm for modern MIL applications~\cite{li2021dual,PDL,shao2021transmil,zhang2022dtfd,qiu2023sc,xiang2022exploring}. Following this line of work, a related but concurrent work~\cite{anonymous2024inherently} attempted to apply ABMIL and its variants to time series. However, ABMIL operates under the assumption that instances are independent and identically distributed, which inherently limits the modeling of the temporal dependencies among instances in time series data.
% dependencies among instances. 
% To address this limitation, most of its follow-up works have been proposed to explicitly model the instance dependencies \cite{campanella2019clinical,tu2019multiple,li2021dual,shao2021transmil,xiang2022exploring}. 

While being related to methods presented in ~\cite{milmtsc,anonymous2024inherently}, our method differs from them in the following ways.  
\textbf{(i)} Unlike the autoregressive hidden Markov used in~\cite{milmtsc}, which struggles with long-range and complex dependencies, the proposed method employs self-attention to model the instance dependencies regardless of their distance, offering inherent interpretability. 
% Furthermore, the proposed method is inherently interpretable (Section~\ref{sec:interp}) and more computationally efficient compared to~\cite{milmtsc}. 
\textbf{(ii)} Authors of \cite{anonymous2024inherently} directly applied MIL to the time series classification problem to obtain interpretability without providing a theoretical justification of how time series classification can be framed as a MIL problem. Especially for the multi-class cases, they violated the MIL assumption that a bag is positive as long as a positive instance is present. Additionally, their proposed method also falls into the category of ABMIL, which does not naturally model the temporal correlation and ordering of time points within a time series. 
In contrast, we re-frame the MIL for more complex multi-class MTSC tasks. Specifically, we show how to effectively tackle multi-class problems in the context of the binary MIL paradigm (Section~\ref{sec:prove}), clarifying and addressing the limitations of standard (AB)MIL techniques used for time series analyses (Section~\ref{sec:time_aware}).

% particularly their inability to account for the temporal ordering and correlation inherent in time series data 

% with RNN~\cite{campanella2019clinical}, GNN~\cite{tu2019multiple}, non-local attention~\cite{li2021dual}, and transformer~\cite{shao2021transmil,xiang2022exploring}. 

% \textbf{(ii)} Although the proposed method also utilizes transformer to model the instance dependencies, it is essentially different from~\cite{xiang2022exploring}. Their studies were applied to histological images, and hence focused on modeling similarities between instances (by self-attention) rather than sequential ordering. Consequently, they did not explicitly apply \textit{positional encoding} to their transformers. However, we prove the importance of \textit{positional encoding} in TimeMIL (see Theorem~\ref{thm:3}). Importantly, we further propose a \textit{wavelet positional encoding} to model sequential ordering with changing frequency over time (see Section~\ref{sec:time_aware}).

% \vspace{-0.1in}
\section{Method}
% \vspace{-0.1in}
In this section, we introduce three key components for applying the proposed TimeMIL to MTSC. First, we formulate the MTSC as a MIL problem in Sec.~\ref{problem} and~\ref{mil-multiplelabel}. Second, we introduce a time-aware MIL pooling to capture the temporal ordering in time series through a wavelet-augmented transformer (Sec.~\ref{sec:time_aware}). Third, we introduce how to quantify the importance of instances in Sec.~\ref{sec:interp}. The entire framework of the proposed TimeMIL is depicted in Fig. \ref{fig:network}. A summary of the proposed framework is presented in Algorithm \ref{alg:1}.

\vspace{-0.1cm}
\subsection{Problem Formulation}
\label{problem}
Multivariate time series data is typically presented as $\{\boldsymbol{X}_1, \cdots, \boldsymbol{X}_n\}$, where $\boldsymbol{X}_i=\left\{\boldsymbol{x}_i^1, \cdots, \boldsymbol{x}_i^T \right\}$ is a time series contains $T$ time points, with each time point $\boldsymbol{x}_i^{t} \in \mathbb{R}^{d}$ being a $d$-dimensional vector. It is noteworthy that the time points are shift-variant and ordered. The goal of MTSC is to learn a direct mapping from feature space $\mathcal{X}$ to label space $\mathcal{Y}$ using the training data $\{(\boldsymbol{X}_1, y_1), \cdots, (\boldsymbol{X}_n, y_n)\}$, where $y_i$ is the label for each time series. 

% $\boldsymbol{X}_i=\left\{\boldsymbol{x}_1, \boldsymbol{x}_2, \cdots, \boldsymbol{x}_T\right\} \in$ $\mathbb{R}^{d \times T}$, where $d$ and $T$ denote the dimensionality and the sequence length, respectively. It is noteworthy that the index $t$ of $x_t\in\mathbb{R}^{d}$ is ordered and is shift-variant. 

% MTSC is simliar to other classification tasks that learning a model from multiple labelled time series and then use $\boldsymbol{X}$ to predict its discrete target or label $\boldsymbol{Y}$. In the following content, we will first introduce the concept of Multiple Instance Learning (MIL) and how it can be generalized to MTSC problem. Accordingly, we will present our proposed framework MIL-WPE.

% \vspace{-0.1cm}
\subsection{MTSC as A MIL Problem} 
\label{mil-multiplelabel}
% \arr{}better to use a MIL. like here $https://www.sciencedirect.com/science/article/pii/S0031320317304065$}
\label{sec:prove}
\noindent\textbf{Binary MTSC.} 
Without violating the MIL assumption, we take binary MTSC as a starting point in the following derivations, then extend it to the multi-class scheme. The goal of a binary MIL is to assign a label to a bag of instances. The MTSC can naturally be formulated as a MIL problem by treating each time series as a bag, with each time slot being an instance. Formally, the binary MTSC under the MIL formulation is defined as 
\begin{equation}
    y_i=\left\{\begin{array}{l}
    0, \text { iff } \sum_{t=1}^{T} y_i^t =0, \ y_i^t \in\{0,1\} \\
    1, \text { otherwise, }
    \end{array}\right.
    \label{eqn:binary_mil}
\end{equation}
where $y_i^t$ denotes the label for each time point indicating if an event of interest has happened at time point $\boldsymbol{x}_i^t$. 
The $\{y_i^1, \cdots, y_i^t\}$ are also known as the instance-level labels in the context of MIL, which are unknown in most scenarios. Eq. \ref{eqn:binary_mil} implies that a time series (bag) $\boldsymbol{X}_i$ is labeled as positive if and only if any of its instance labels is positive, negative otherwise. 
% \arr{I think the next phrase is obvious and can be discarded because $y_i^t$ is the label of $x_i^t$ by definition} Accordingly, an instance $\boldsymbol{x}_i^t$ is positive if and only if its corresponding instance label $y_i^t$ is 1; and negative otherwise. 

The bag-level prediction $\hat{y}_i$ of a MIL is given as a score function $S: \mathcal{X} \rightarrow \mathbb{R}$~\cite{ilse2018attention}: 
\begin{equation}
    \hat{y}_i = S(\boldsymbol{X}_i),
\end{equation}
where the outcome of a score function is a probability.

\begin{theorem}\label{thm:1}
\cite{ilse2018attention,shao2021transmil} Suppose the score function $S$ is a $(\delta_{\varepsilon},\varepsilon)$-continuous symmetric function w.r.t Hausdorff distance $d_H(\cdot, \cdot)$, i.e. $\forall d_H(\boldsymbol{X}_i, \boldsymbol{X}_j)<\delta_{\varepsilon}$, we have $ |S(\boldsymbol{X}_i)-S(\boldsymbol{X}_j)|< \varepsilon$, for
 $\forall \varepsilon>0$. For any invertible map $\sigma: \mathcal{X} \rightarrow \mathbb{R}^d$, $S$ can be approximated by certain continuous functions $g$ and $f$:
\begin{align}
|S(\boldsymbol{X}_i)-g(\sigma\{f(\boldsymbol{x}_i^t): \boldsymbol{x}_i^t \in \boldsymbol{X}_i\})|<\varepsilon .
\end{align}
\end{theorem}

Theorem~\ref{thm:1} defines the generic pipeline of a MIL, which consists of three main parts: (i) The function $f$ is a feature extractor that projects the input instances into $L$-dimensional vector embeddings $\Tilde{\boldsymbol{X}}_i$. (ii) $\sigma$ is known as the MIL pooling function that aggregates instance vector embeddings into a single vector. It should be noted that the original MIL pooling function $\sigma$ should be permutation-invariant~\cite{ilse2018attention}.
(iii) $g$ denotes the bag-level classifier (e.g., a linear classifier) that maps the vector embedding after applying MIL pooling to a bag-level probability prediction $\hat{y}_i \in [0, 1]$. 

\noindent\textbf{Mutli-Class MTSC.} 
A multi-class time series classification with a total of $C$ classes can be performed as several \textit{one-vs-rest} binary MIL without violating its assumption:
\begin{align}
y_{i, c} =\left\{\begin{array}{l}
0, \text { iff } \sum_{t=1}^{T} y_{i,c}^t=0, \ y^{t}_{i,c} \in\{0,1\} \\
1, \text { otherwise,}
\end{array}\right.
\label{eqn:multi_mil}
\end{align}
where $y_{i, c}^t=1$ denotes a time point with significant contribution to class $c\in \{1,\cdots,C\}$. The final bag-level prediction $y_i$ for a bag is computed as the class with the highest probability: 
\begin{equation}
    \hat{y}_i = \operatorname{argmax}\limits_{c} \ \hat{y}_{i, c},
\end{equation}
which is consistent with the one-vs-rest scheme.
% \xc{which is consistent with the conventional one-vs-rest scheme.}
% meaning that a bag is positive (i.e., $y_{i, c}=1$) as long as a positive instance related to a certain category $c$ is present.
% \arr{Is this the case, even if instances are multi-class, we assign binary 0-1 labels to the bag of samples? It sounds counterintuitive}
% This is also known as 

\begin{remark}\label{remark:1}
    The MIL in Theorem~\ref{thm:1} fails to model the temporal ordering among time points within a time series.
\end{remark}

Remark~\ref{remark:1} arises from the symmetric (permutation-invariant) property of the MIL pooling function $\sigma$. The function $\sigma$ remains the same for every permutation of the instances within a bag, thereby neglecting the temporal ordering between time points (instances) in time series modeling. This hinders the direct translation of classic MIL into MTSC tasks. 
To address this limitation, we propose a time-aware MIL pooling in Sec.~\ref{sec:time_aware}. 

\begin{algorithm}[!t]
% \scriptsize
% \label{alg:1}
\caption{Time-aware MIL (Forward Propagation) } 
\begin{algorithmic}[1]\label{alg:1}
\REQUIRE Input sequence $\boldsymbol{X}_i=\left\{\boldsymbol{x}_i^1, \cdots, \boldsymbol{x}_i^T \right\}$. A neural network contains:  $f$: feature extractor, $g$: bag-level classifier, $\operatorname{WPE}$ bases, and class token $\boldsymbol{x}^{\operatorname{cls}}$.
 \ENSURE Predicted label $\hat{y}_i$.
\STATE Get embedding: $\left\{\boldsymbol{x}_i^1, \cdots, \boldsymbol{x}_i^T \right\}\leftarrow f(\left\{\boldsymbol{x}_i^1, \cdots, \boldsymbol{x}_i^T \right\})$ 

\textcolor{gray}{\#MIL pooling from here (Section 3.3)}
\STATE Init class token for the bag: $\boldsymbol{x}_i^{\operatorname{cls}}\leftarrow \boldsymbol{x}^{\operatorname{cls}}$.
\STATE Add class token: $\boldsymbol{X}_i^{\operatorname{cls}} = \boldsymbol{x}_i^{\operatorname{cls}}\cup \boldsymbol{X}_i$.
\FOR{j=1:2}
\STATE PE: $\boldsymbol{X}_i\leftarrow \boldsymbol{X}_i+WPE_j(\boldsymbol{X}_i)$. \textcolor{gray}{\# $WPE_j(\boldsymbol{X}_i)$ is from Eq. 10. Here, $\boldsymbol{X}_i$ is the part of $\boldsymbol{X}_i^{\operatorname{cls}}$. } 
\STATE $\boldsymbol{X}_i^{\operatorname{cls}}\leftarrow Transformer_j(\boldsymbol{X}_i^{\operatorname{cls}})$.
\ENDFOR
\\ \textcolor{gray}{\#Bag-level classification from here}
\STATE $\hat{y}_i\leftarrow g(\boldsymbol{x}_i^{\operatorname{cls}})$. \textcolor{gray}{\#$\boldsymbol{x}_i^{\operatorname{cls}}$ is obtained from $\boldsymbol{X}_i^{\operatorname{cls}}$.}

% Thank you for the reminder. In the revised version, the supplementary section will feature more comprehensive descriptions and analyses of the ablation study. 

% \textcolor{blue}{We show the forward propagation of our MIL framework. In the training phase, each part (i.e. $f$, $g$, parameters in $WPE$, and $\boldsymbol{x}^{\operatorname{cls}}$) can be updated by the regular gradient-based optimization. }
\end{algorithmic}
\end{algorithm}
% \vspace{-0.1in}
\subsection{Time-Aware MIL Pooling for MTSC}
\label{sec:time_aware}
% \vspace{-0.1in}
From an entropy perspective, understanding permutation-variant properties in time series can be quite insightful. As discussed in Sec.~\ref{sec:prove}, we assume that each time point (instance) $\boldsymbol{x}_i^j$ in a time series (bag) is a realization of a random variable $\Theta^j$ conditioned by a time index $t^j$. The resulting bag can be represented as a random variable $\boldsymbol{X} = \{(\Theta^1|t^1), \cdots, (\Theta^T|t^T)\}$, where $t^{j}\neq t^{k},\forall j\neq k$. Likewise, $\boldsymbol{Y}$ denotes a random variable for bag label. Here, we use notation $\Theta^j$ for an instance to distinguish a random variable (often as uppercase) and its realization (often as lowercase).  It is fair to construct a general assumption:  
$p(\Theta^{j}|t^{j})\neq p(\Theta^{j}|t^{k})$, which indicates an instance varies when it is presented in different locations. 

\begin{proposition}\label{prop:2}
%\pq{We would recommend being more direct, saying that the arbitrary permutation increases the uncertainty in terms of block entropy. OTherw}
 % For a sequence where the correct knowledge representation necessitates the modeling of permutation information.
Shuffling the time points within a time series potentially disrupts its predictability. This means, under the general assumption, the entropy before and after shuffling typically differs, i.e., the equality:
 % \begin{align}\nonumber
 % H( \cdots, (\Theta^j|t^j),\cdots) = \texttt{shuffle}(H(\cdots, (\Theta^j|t^{j}),\cdots)),   
 % \end{align}
 \begin{align}\nonumber
 H( \cdots, (\Theta^j|t^j),\cdots) =H(\cdots, (\Theta^j|t^{\Bar{j}}),\cdots),   
 \end{align} 
does \textbf{not} always hold. Here, $t^{\Bar{1}}, \cdots,t^{\Bar{T}}$ are sampled from the set $\{t^1,\cdots,t^T\}$ without replacement. The right term denotes the time series after randomly shuffling. 
% Here, $\texttt{shuffle}$ denotes an arbitrary permutation of time points within a time series.
\end{proposition}
% \begin{proof}
   
% \end{proof}

\begin{figure}[t]
\centering\includegraphics[width=0.4\textwidth]{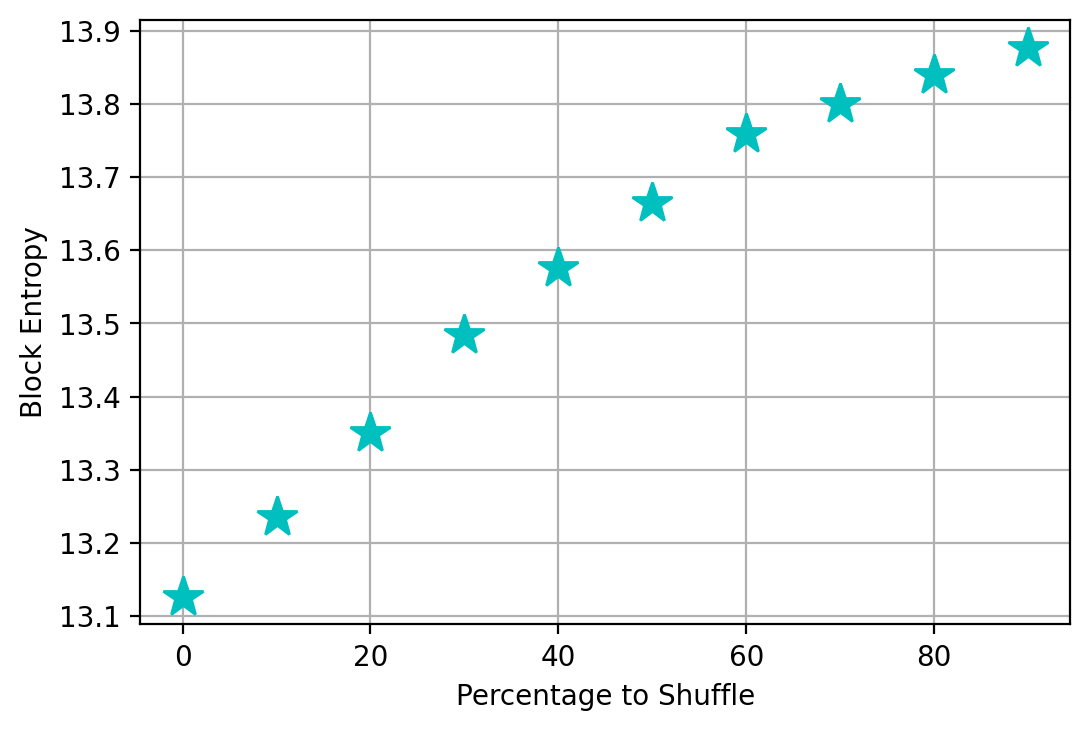}
\vspace{-0.2cm}
\caption{The block entropy in Shakespeare's Sonnets with varying shuffling rates, where the higher shuffling rates result in higher block entropy.} 
\vspace{-0.2cm}
\label{fig:blockentropy}
\end{figure}
% \vspace{-0.1in}
Please refer to Appendix \ref{appendix:propproof} for its proof. Proposition~\ref{prop:2} implies that random permutation of time points within a regular time series potentially increases its uncertainty in terms of entropy. Since it is challenging to directly compute the entropy for high-dimensional continuous-valued variables from the observation, we illustrate this fact by using a supportive example in text sequence (Shakespeare's Sonnets\footnote{ \url{https://shakespeares-sonnets.com/Archive/allsonn.htm}.}). The feasible domain of the text sequence is a discrete 1D domain, and its uncertainty can be measured directly by the Shannon block entropy \cite{shannon1951prediction}. Please refer to Appendix \ref{appendix:be} for more detail about Shannon block entropy. As shown in Fig. \ref{fig:blockentropy}, the block entropy increases as the shuffling rate increases. We will demonstrate the importance of not just modeling the random permutation but also accounting for the temporal correlation between instances.

% In fact, neglecting the sequential information often increases the complexity and disrupt the predictability of a regular time series, since a regular time series often demonstrates semantically meaningful patterns. Since it is challenging to compute the entropy purely for high-dimensional continuous-valued variables from the observation, we show the fact by using an example of a text sequence (Shakespeare's Sonnets\footnote{\url{https://shakespeares-sonnets.com/Archive/allsonn.htm}.}), whose feasible domain is 1D discrete and its uncertainty is able to be measured by Shannon block entropy \cite{shannon1951prediction}. We randomly shuffle a certain ratio of the text. Fig. \ref{fig:blockentropy} demonstrates random permutation increases consistently increase the uncertainty of the sequence.

\begin{theorem}\label{thm:3} \nonumber
    Modeling the temporal correlation between instances lowers the complexity of developing a good classifier, which is presented by class-conditioned entropy:
    % \begin{align}
    %      H((\Theta^1|t^1), \cdots, (\Theta^T|t^T)|\boldsymbol{Y}) \leq \underbrace{\sum_{i} H((\Theta^i|t^i) | \boldsymbol{Y})}_{\delta},
    %    \vspace{-3.0cm}
    % \end{align}
\begin{align}
         H(\boldsymbol{X}_{c}|\boldsymbol{Y})\leq H(\boldsymbol{X}_{nc}|\boldsymbol{Y}),
       % \vspace{-3.0cm}
    \end{align}  
    where $H(\boldsymbol{X}_{c}|\boldsymbol{Y})=H((\Theta^1|t^1), \cdots, (\Theta^T|t^T)|\boldsymbol{Y})$ is derived under modeling the correlation among instances within a bag while $H(\boldsymbol{X}_{nc}|\boldsymbol{Y})=\sum_{i} H((\Theta^i|t^i) | \boldsymbol{Y})$ is derived under the assumption that instances are independent and identically distributed.
% where the term $\delta$ is derived under the assumption that instances are independent and identically distributed.
\end{theorem}
\vspace{-0.1in}
\begin{proof}
For convenience, we denote $(\Theta^j|t^j)$ by $\Lambda^j$. 
    \begin{equation}
    \begin{split}
        % H(\boldsymbol{X}|\boldsymbol{Y}) &= H(\Lambda^1, \cdots, \Lambda^T|\boldsymbol{Y}) \\
        % H(\Lambda^1,  \boldsymbol{Y}) &= H( \boldsymbol{Y}) + H(\Lambda^1 | \boldsymbol{Y})\\
        % H(\Lambda^1, \Lambda^2,  \boldsymbol{Y}) &= H( \boldsymbol{Y}) + H(\Lambda^1 | \boldsymbol{Y})+ H(\Lambda^2 | \Lambda^1,\boldsymbol{Y})\\
        % &\cdots \\
        H(\Lambda^1, \cdots,\Lambda^T |\boldsymbol{Y}) &= H(\Lambda^T| \Lambda^{T-1}, \cdots, \Lambda^1,\boldsymbol{Y}) \\ & + H(\Lambda^{T-1} | \Lambda^{T-2}, \cdots, \Lambda^{1}, \boldsymbol{Y})\\  &+ \cdots\\ 
        &+ H(\Lambda^1 | \boldsymbol{Y}) %\\
        % &+ H(\boldsymbol{Y})%
        % \\
        % & \leq \sum_{i} H(\Lambda^i )
    \end{split}
    \end{equation}
Since $H(\Lambda^{i} | \Lambda^{i-1}, \cdots, \Lambda^{1}, \boldsymbol{Y})\leq H(\Lambda^{i}|\boldsymbol{Y})$, which indicates knowing more information lowers the uncertainty:
    \begin{align}
        H(\Lambda^1, \cdots,\Lambda^T |\boldsymbol{Y})  \leq \sum_{i} H(\Lambda^i | \boldsymbol{Y})
    \end{align}
 \vspace{-0.2cm}
\end{proof}
   
\begin{remark}
    The conditional entropy $H(\boldsymbol{X}|\boldsymbol{Y})$
 measures the uncertainty of the bag feature $\boldsymbol{X}$ given that the bag-level class label $\boldsymbol{Y}$ is known. In the context of classification, it quantifies the spread of features within each class. A high value of $H(\boldsymbol{X}|\boldsymbol{Y})$ indicates that the features belonging to the same class can vary significantly. This suggests that the features are not clustered tightly but are spread out. In contrast, a lower $H(\boldsymbol{X}|\boldsymbol{Y})$ suggests that the features from the same class are more homogeneous, exhibiting less variability. This homogeneity can make it easier to classify instances since the features within each class are more consistent, potentially resulting in a simpler decision boundary.
\end{remark}

% \vspace{-0.05in}
Theorem~\ref{thm:3} immediately implies the benefit of modeling temporal permutation and correlation between instances and provides a generic formulation of time-aware MIL pooling. The realization of Theorem~\ref{thm:3} can be achieved by a transformer with the unique token mechanism.
% The realization of Theorem~\ref{thm:3} can vary from traditional auto-regressive models to recurrent neural networks, and even to transformers. We opt for the transformer in this paper due to its excellent scaling behaviors and unique token mechanism.
First, transformers help tackle the conditional entropy $H(\Lambda^{t}| \Lambda^{t-1}, \cdots, \Lambda^{1},\boldsymbol{Y}) $ in Theorem~\ref{thm:3} by employing a \textit{class token} $\boldsymbol{x}_i^{\operatorname{cls}}$. The yielded tokenized bag of instances is $\boldsymbol{X}_i^{\operatorname{cls}} = \{\boldsymbol{x}_i^{\operatorname{cls}}, \boldsymbol{x}_i^1, \cdots, \boldsymbol{x}_i^T\}$.  Second, we propose a novel \textit{positional encoding} in our transformer-based pooling through the lens of wavelet theory to further capture the multi-scale time-frequency ordering relationship among instances.
% It is noteworthy that after the transformer blocks, only the class token is fed into the final classifier.

% \subsubsection{Temporal Ordering as Self-Attention}
\noindent\textbf{Temporal correlation as self-Attention.}
The self-attention mechanism~\cite{selfattention} is proposed to capture mutual information between time points. In the context of MIL, we use multi-head self-attention (MHSA) to model the sequential correlation between instances:
\begin{align}
     \operatorname{MHSA}(\boldsymbol{X}_i^{\operatorname{cls}}) = [\operatorname{head}_1, \cdots, \operatorname{head}_H]\boldsymbol{W}_0
     \vspace{-0.6cm}
\end{align}
where:
\vspace{-0.2cm}
\begin{align}\label{eqn:self_attn}
    \nonumber&\operatorname{head}_h = \operatorname{Attention}(\boldsymbol{X}_i^{\operatorname{cls}}\boldsymbol{W}_h^Q, \boldsymbol{X}_i^{\operatorname{cls}} \boldsymbol{W}_h^K, \boldsymbol{X}_i^{\operatorname{cls}}\boldsymbol{W}_h^V) \\
     &\operatorname{Attention} (\boldsymbol{Q}, \boldsymbol{K}, \boldsymbol{V}) = \operatorname{softmax}(\boldsymbol{Q}\boldsymbol{K}^{\mathbf{T}}/\sqrt{d_k})\boldsymbol{V},
\end{align}
where $\boldsymbol{W}_0$, $\boldsymbol{W}_h^Q$, $\boldsymbol{W}_h^K$, $\boldsymbol{W}_h^V$ are trainable parameters. It is noteworthy that after the transformer blocks, we only pass the \textit{class token} to the bag-level classifier $g$ to make a prediction.
However, standard self-attention has quadratic time and memory complexity $\mathcal{O}(T^2)$ w.r.t. the number of instances in a bag ($T$). Recent advances~\cite{nys,wang2020linformer,shen2021efficient} in self-attention studies have reduced the quadratic complexity to approximately linear. Specifically, we use the approximation of self-attention proposed by~\cite{nys} in our implementation to reduce the complexity of the proposed TimeMIL.

% \subsubsection{Wavelet Positional Encoding}\label{sec:WPE}
\noindent\textbf{Wavelet positional encoding.}
The classic transformers use \textit{Sinusoidal} positional encoding to capture the relative ordering in a time series, as self-attention does not take the temporal ordering of time points into account. The disadvantage of \textit{Sinusoidal} positional encoding is that it is pre-defined and non-learnable. Importantly, the \textit{Sinusoidal} positional encoding is independently generated away from the input context; hence, it cannot capture both the time and frequency information of time series. To address this limitation, we propose a learnable \textit{wavelet} positional encoding in a conditional positional encoding fashion~\cite{chen2023rethinking,chu2023conditional}: 
\vspace{-0.2cm}
\begin{equation}\label{eq:wpe_1}
\vspace{-0.2cm}
    \operatorname{WPE}(\boldsymbol{X}_i) = \sum_{j=1}^{n_{\operatorname{W}}} \Phi(\boldsymbol{X}_i,\boldsymbol{\Psi}_j),
    \vspace{-0.1cm}
\end{equation}
where $n_{\operatorname{W}}$ denotes the number of wavelet basis, which is empirically set to 3 in this paper.  $\{\boldsymbol{\Psi}_1,\cdots,\boldsymbol{\Psi}_{n_{\operatorname{W}}}\}$ are learnable wavelet kernels with $\boldsymbol{\Psi}_j=\{\psi_{a_{j1}, b_{j1}}(t),\cdots, \psi_{a_{jL}, b_{jL}}(t)\}$ where $\psi$ is \textit{mother wavelet}, which is chosen to be the Mexican hat in our experiments.

The \textit{Gabor-Heisenberg limit} \cite{gabor1946theory}, which is the uncertainty principle in the time-frequency version, states that it is impossible to precisely determine both the time and frequency of a signal, simultaneously. In the context of the wavelet transform, this principle implies a trade-off: the higher the resolution needed in time, the lower the resolution becomes in frequency. This implies that the careful selection of time-frequency resolution is crucial for effectively characterizing different signals. Hence, we learn the scaling and translation parameters $\{(a_{j1}, b_{j1}), \cdots, (a_{jL}, b_{jL}) \}$ to form the wavelet basis from \textit{mother wavelet}. 
$\Phi(\cdot)$ is the channel-wise wavelet transform, which can be formulated as convolving the input signal with the wavelet kernels:
\begin{align}\label{eq:wpe_2}
    \Phi(\Tilde{\boldsymbol{X}}_i,\boldsymbol{\Psi}_j) = \left[\begin{array}{c}
\Tilde{\boldsymbol{X}}_{i1} \circledast  \psi_{a_{j1}, b_{j1}} \\
% \Tilde{\boldsymbol{X}}_{i2} \circledast  \psi_{a_{j2}, b_{j2}} \\
\vdots \\
\Tilde{\boldsymbol{X}}_{iL} \circledast  \psi_{a_{jL}, b_{jL}} 
\end{array}\right]^\top \in \mathbb{R}^{n\times L}.
 \end{align}

\begin{figure}[!t]
    \centering
    \includegraphics[width=0.48\textwidth]{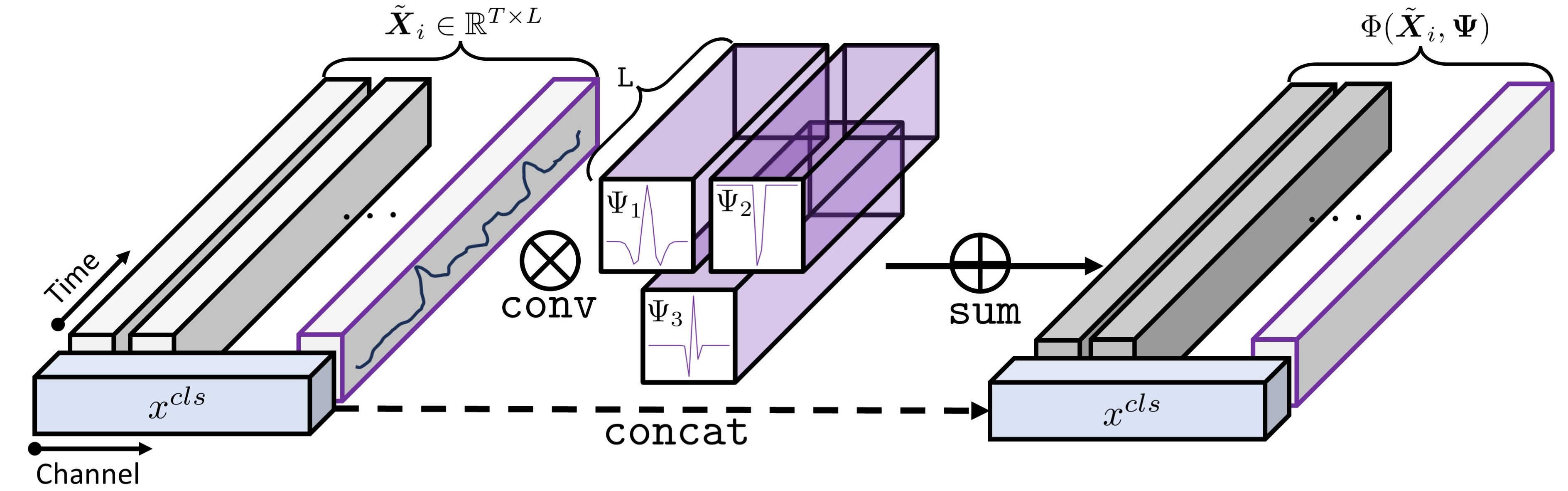}
    \vspace{-0.6cm}
    \caption{The proposed learnable wavelet positional encoding: First, wavelet transform is performed for the input signal (by excluding the class token) with each wavelet basis (Eq. \ref{eq:wpe_2}). Second, the signals are aggregated in the wavelet domain by a summation (Eq. \ref{eq:wpe_1}). In the case of $n_w=3$, we use 3 learnable wavelet bases ($\boldsymbol{\Psi}_1,\boldsymbol{\Psi}_2,\boldsymbol{\Psi}_3$) to model changing frequency and time scales.}  
    \vspace{-0.2cm}
    \label{fig:wpe}
\end{figure}

The resulting WPE is depicted in Fig.~\ref{fig:wpe}.

\subsection{Interpretability}\label{sec:interp}
The proposed time-aware MIL is naturally interpretable due to its ability to localize time points of interest within a time series. One way to achieve this is to hack into the attention map of the transformer layers in the proposed time-aware MIL pooling. The attention map (refer to Eq. \ref{eqn:self_attn}) measures the importance of each instance $\boldsymbol{x}_i^t$ in the series $\boldsymbol{X}_i$ in the MIL pooling:
% \begin{equation}
%     \alpha_i^t = \operatorname{softmax}\left(\frac{\boldsymbol{x}_i^{\operatorname{cls}} \boldsymbol{W}_h^Q \ \boldsymbol{x}_i^t \boldsymbol{W}_h^K }{\sqrt{d_k}} \right),
% \end{equation}
\begin{equation}\label{eqn:attn}
    \boldsymbol{A}_i = \operatorname{softmax}\left(\frac{\boldsymbol{x}_i^{\operatorname{cls}} \boldsymbol{W}_h^Q \ \boldsymbol{X}_i\boldsymbol{W}_h^K }{\sqrt{d_k}} \right),
\end{equation}

where $\operatorname{softmax}$ is performed over the time dimension.
$\boldsymbol{A}_i\in\mathbb{R}^T$ is the importance weights of all instances, with its $t$-th element corresponding to the importance of the time point $\boldsymbol{x}_i^t$. We only use the \textit{class token} to calculate the importance weight, as it determines the most relevant time points that characterize a certain class label in MTSC tasks.

\begin{table*}[ht]
\centering
\caption{Results of Group 1 Experiments. Comparison with the recent state-of-the-art MTSC methods on 26 datasets. The best results are highlighted by {\color[HTML]{FF0000} \textbf{bold}} and the second best are highlighted by \underline{\color[HTML]{0000FF} underline}. 'N/A' in the table denotes the corresponding method was unable to obtain results due to memory or computational limitations \cite{liu2023todynet}.}
\label{tab:result1}
\resizebox{1.0\textwidth}{!}{%
\begin{tabular}{l|ccccccccccc}\toprule
\multicolumn{1}{c}{\textbf{Datasets/Methods}} & \textbf{ED-1NN} & \textbf{DTW-1NN-I} & \textbf{DTW-1NN-D} & \textbf{\begin{tabular}[c]{@{}c@{}}MLSTM-FCN \\      \textit{Neur. Net.'19}\end{tabular}} & \textbf{\begin{tabular}[c]{@{}c@{}}ShapeNet \\      \textit{AAAI'21}\end{tabular}} & \textbf{\begin{tabular}[c]{@{}c@{}}WEASEL+MUSE\\\textit{arxiv'2017}\end{tabular}} & \textbf{\begin{tabular}[c]{@{}c@{}}TapNet \\      \textit{AAAI'20}\end{tabular}} & \textbf{\begin{tabular}[c]{@{}c@{}}OS-CNN\\       \textit{ICLR'22}\end{tabular}} & \textbf{\begin{tabular}[c]{@{}c@{}}MOS-CNN \\      \textit{ICLR'22}\end{tabular}} & \textbf{\begin{tabular}[c]{@{}c@{}}TodyNet \\      \textit{arxiv'23}\end{tabular}} & \textbf{Ours} \\ \midrule
ArticularyWordRecognition & 0.970 & 0.980 & 0.987 & 0.973 & 0.987 & \underline{\color[HTML]{0000FF} 0.990} & 0.987 & 0.988 & {\color[HTML]{FF0000} \textbf{0.991}} & 0.987 & \underline{\color[HTML]{0000FF} 0.990} \\
AtrialFibrillation & 0.267 & 0.267 & 0.200 & 0.267 & 0.400 & 0.333 & 0.333 & 0.233 & 0.183 & {\color[HTML]{0000FF} 0.467} & {\color[HTML]{FF0000} \textbf{0.733}} \\
BasicMotions & 0.675 & {\color[HTML]{FF0000} \textbf{1.000}} & \underline{\color[HTML]{0000FF} 0.975} & 0.950 & {\color[HTML]{FF0000} \textbf{1.000}} & {\color[HTML]{FF0000} \textbf{1.000}} & {\color[HTML]{FF0000} \textbf{1.000}} & {\color[HTML]{FF0000} \textbf{1.000}} & {\color[HTML]{FF0000} \textbf{1.000}} & {\color[HTML]{FF0000} \textbf{1.000}} & {\color[HTML]{FF0000} \textbf{1.000}} \\
Cricket & 0.944 & 0.986 & {\color[HTML]{FF0000} \textbf{1.000}} & 0.917 & 0.986 & {\color[HTML]{FF0000} \textbf{1.000}} & 0.958 & \underline{\color[HTML]{0000FF} 0.993} & 0.990 & {\color[HTML]{FF0000} \textbf{1.000}} & {\color[HTML]{FF0000} \textbf{1.000}} \\
DuckDuckGeese & 0.275 & 0.550 & 0.600 & 0.675 & \underline{\color[HTML]{0000FF} 0.725} & 0.575 & 0.575 & 0.540 & 0.615 & 0.580 & {\color[HTML]{FF0000} \textbf{0.780}} \\
EigenWorms & 0.550 & 0.603 & 0.618 & 0.504 & \underline{\color[HTML]{0000FF} 0.878} & {\color[HTML]{FF0000} \textbf{0.890}} & 0.489 & 0.414 & 0.508 & 0.840 & 0.823 \\
Epilepsy & 0.667 & 0.978 & 0.964 & 0.761 & 0.987 & {\color[HTML]{FF0000} \textbf{1.000}} & 0.971 & 0.980 & 0.996 & 0.971 & {\color[HTML]{FF0000} \textbf{1.000}} \\
EthanolConcentration & 0.293 & 0.304 & 0.323 & 0.373 & 0.312 & 0.133 & 0.323 & 0.240 & {\color[HTML]{FF0000} \textbf{0.415}} & 0.350 & \underline{\color[HTML]{0000FF} 0.407} \\
ERing & 0.133 & 0.133 & 0.133 & 0.133 & 0.133 & 0.430 & 0.133 & 0.881 & \underline{\color[HTML]{0000FF} 0.915} & {\color[HTML]{0000FF} 0.915} & {\color[HTML]{FF0000} \textbf{0.956}} \\
FaceDetection & 0.519 & 0.513 & 0.529 & 0.545 & 0.602 & 0.545 & 0.556 & 0.575 & 0.597 & \underline{\color[HTML]{0000FF} 0.627} & {\color[HTML]{FF0000} \textbf{0.698}} \\
FingerMovements & 0.550 & 0.520 & 0.530 & 0.580 & \underline{\color[HTML]{0000FF} 0.589} & 0.490 & 0.530 & 0.568 & 0.568 & 0.570 & {\color[HTML]{FF0000} \textbf{0.670}} \\
HandMovementDirection & 0.279 & 0.306 & 0.231 & 0.365 & 0.338 & 0.365 & 0.378 & 0.443 & 0.361 & {\color[HTML]{FF0000} \textbf{0.649}} & \underline{\color[HTML]{0000FF} 0.487} \\
Handwriting & 0.371 & 0.509 & 0.607 & 0.286 & 0.451 & 0.605 & 0.357 & \underline{\color[HTML]{0000FF} 0.668} & {\color[HTML]{FF0000} \textbf{0.677}} & 0.436 & 0.482 \\
Heartbeat & 0.620 & 0.659 & 0.717 & 0.663 & \underline{\color[HTML]{0000FF} 0.756} & 0.727 & 0.751 & 0.489 & 0.604 & \underline{\color[HTML]{0000FF} 0.756} & {\color[HTML]{FF0000} \textbf{0.815}} \\
Libras & 0.833 & 0.894 & 0.872 & 0.856 & 0.856 & 0.878 & 0.850 & 0.950 & \underline{\color[HTML]{0000FF} 0.965} & 0.850 & {\color[HTML]{FF0000} \textbf{0.972}} \\
LSST & 0.456 & 0.575 & 0.551 & 0.373 & 0.590 & 0.590 & 0.568 & 0.413 & 0.521 & \underline{\color[HTML]{0000FF} 0.615} & {\color[HTML]{FF0000} \textbf{0.690}} \\
MotorImagery & 0.510 & 0.390 & 0.500 & 0.510 & 0.610 & 0.500 & 0.590 & 0.535 & 0.515 & \underline{\color[HTML]{0000FF} 0.640} & {\color[HTML]{FF0000} \textbf{0.720}} \\
NATOPS & 0.860 & 0.850 & 0.883 & 0.889 & 0.883 & 0.870 & 0.939 & 0.968 & 0.951 & \underline{\color[HTML]{0000FF} 0.972} & {\color[HTML]{FF0000} \textbf{0.994}} \\
PenDigits & 0.973 & 0.939 & 0.977 & 0.978 & 0.977 & 0.948 & 0.980 & \underline{\color[HTML]{0000FF} 0.985} & 0.983 & {\color[HTML]{FF0000} \textbf{0.987}} & 0.600 \\
PEMS-SF & 0.705 & 0.734 & 0.711 & 0.699 & 0.751 & N/A  & 0.751 & 0.760 & 0.764 & \underline{\color[HTML]{0000FF} 0.780} & {\color[HTML]{FF0000} \textbf{0.931}} \\
PhonemeSpectra & 0.104 & 0.151 & 0.151 & 0.110 & 0.298 & 0.190 & 0.175 & 0.299 & 0.295 & {\color[HTML]{0000FF} 0.309} & {\color[HTML]{FF0000} \textbf{0.311}} \\
RacketSports & 0.868 & 0.842 & 0.803 & 0.803 & 0.882 & {\color[HTML]{FF0000} \textbf{0.934}} & 0.868 & 0.877 & \underline{\color[HTML]{0000FF} 0.929} & 0.803 & 0.908 \\
SelfRegulationSCP1 & 0.771 & 0.765 & 0.775 & 0.874 & 0.782 & 0.710 & 0.652 & 0.835 & 0.829 & {\color[HTML]{FF0000} \textbf{0.898}} & {\color[HTML]{FF0000} \textbf{0.898}} \\
SelfRegulationSCP2 & 0.483 & 0.533 & 0.539 & 0.472 & \underline{\color[HTML]{0000FF} 0.578} & 0.460 & 0.550 & 0.532 & 0.510 & 0.550 & {\color[HTML]{FF0000} \textbf{0.639}} \\
StandWalkJump & 0.200 & 0.333 & 0.200 & 0.067 & \underline{\color[HTML]{0000FF} 0.533} & 0.333 & 0.400 & 0.383 & 0.383 & 0.467 & {\color[HTML]{FF0000} \textbf{0.733}} \\
UWaveGestureLibrary & 0.881 & 0.869 & 0.903 & 0.891 & 0.906 & 0.916 & 0.894 & {\color[HTML]{FF0000} \textbf{0.927}} & \underline{\color[HTML]{0000FF} 0.926} & 0.850 & 0.900 \\ \midrule
Ours 1-to-1-Wins & 25 & 23 & 22 & 25 & 22 & 16 & 24 & 22 & 19 & 20 & - \\
Ours 1-to-1-Draws & 0 & 1 & 1 & 0 & 1 & 4 & 1 & 1 & 1 & 3 & - \\
Ours 1-to-1-Losses & 1 & 2 & 3 & 1 & 3 & 5 & 1 & 3 & 6 & 3 & - \\ \midrule
Average accuracy ($\uparrow$) & 0.568 & 0.622 & 0.626 & 0.597 & 0.684 & 0.656 & 0.637 & 0.672 & 0.692 & \underline{\color[HTML]{0000FF} 0.726} & {\color[HTML]{FF0000} \textbf{0.774}} \\
Total best accuracy ($\uparrow$) & 0 & 1 & 1 & 0 & 1 & \underline{\color[HTML]{0000FF} 5} & 1 & 2 & 4 & \underline{\color[HTML]{0000FF} 5} & {\color[HTML]{FF0000} \textbf{18}} \\
Average Rank ($\downarrow$) & 9.154 & 7.904 & 7.231 & 7.865 & 4.731 & 5.900 & 6.500 & 5.442 & 4.692 & \underline{\color[HTML]{0000FF} 4.058} & {\color[HTML]{FF0000} \textbf{2.327}} \\ \bottomrule
\end{tabular}%
}
\end{table*}
% \input{icml2023/table2}

% \usepackage{graphicx}

\begin{table*}[!t]
\centering
% \vspace{-0.2in}
\caption{Results of Group 2 Experiments. Comparison with the recent state-of-the-art general time analysis frameworks on 10 datasets. }
\label{tab:result2}
\resizebox{\textwidth}{!}{%
\begin{tabular}{l|ccccccccccccccccc} \toprule
\multicolumn{1}{c}{\textbf{Datasets/Methods}} & \textbf{\begin{tabular}[c]{@{}c@{}}LSTNet\\      \textit{SIGIR'18}\end{tabular}} & \textbf{\begin{tabular}[c]{@{}c@{}}LSSL\\      \textit{ICLR'22}\end{tabular}} & \textbf{\begin{tabular}[c]{@{}c@{}}Rocket\\      \textit{DMKD'18}\end{tabular}} & \textbf{\begin{tabular}[c]{@{}c@{}}SCINet\\      \textit{NeurIPS'22}\end{tabular}} & \textbf{\begin{tabular}[c]{@{}c@{}}MICN\\      \textit{ICLR'23}\end{tabular}} & \textbf{\begin{tabular}[c]{@{}c@{}}TimesNet\\      \textit{ICLR'23}\end{tabular}} & \textbf{\begin{tabular}[c]{@{}c@{}}Dlinear\\      \textit{AAAI'23}\end{tabular}} & \textbf{\begin{tabular}[c]{@{}c@{}}LightTS\\      \textit{arxiv'22}\end{tabular}} & \textbf{\begin{tabular}[c]{@{}c@{}}MTS-Mixer\\      \textit{arxiv'23}\end{tabular}} & \textbf{\begin{tabular}[c]{@{}c@{}}Rlinear\\      \textit{arxiv'23}\end{tabular}} & \textbf{\begin{tabular}[c]{@{}c@{}}RMLP\\      \textit{arxiv'23}\end{tabular}} & \textbf{\begin{tabular}[c]{@{}c@{}}FEDformer\\      \textit{ICML'22}\end{tabular}} & \textbf{\begin{tabular}[c]{@{}c@{}}Flowformer\\      \textit{ICML'22}\end{tabular}} & \textbf{\begin{tabular}[c]{@{}c@{}}Crossformer\\      \textit{ICLR'23}\end{tabular}} & \textbf{\begin{tabular}[c]{@{}c@{}}PatchTST\\      \textit{ICLR'23}\end{tabular}} & \textbf{\begin{tabular}[c]{@{}c@{}}ModernTCN\\      \textit{ICLR'24}\end{tabular}} & \textbf{Ours} \\ \midrule
EthanolConcentration & 0.399 & 0.311 & {\color[HTML]{FF0000} \textbf{0.452}} & 0.344 & 0.353 & 0.357 & 0.362 & 0.297 & 0.338 & 0.289 & 0.313 & 0.312 & 0.338 & 0.380 & 0.328 & 0.363 & \underline{\color[HTML]{0000FF} 0.407} \\
FaceDetection & 0.657 & 0.667 & 0.647 & 0.689 & 0.652 & 0.686 & 0.680 & 0.675 & \underline{\color[HTML]{0000FF} 0.702} & 0.656 & 0.673 & 0.660 & 0.676 & 0.687 & 0.683 & {\color[HTML]{FF0000} \textbf{0.708}} & 0.698 \\
Handwriting & 0.258 & 0.246 & {\color[HTML]{FF0000} \textbf{0.588}} & 0.236 & 0.255 & 0.321 & 0.270 & 0.261 & 0.260 & 0.281 & 0.300 & 0.280 & 0.338 & 0.288 & 0.296 & 0.306 & \underline{\color[HTML]{0000FF} 0.487} \\
Heartbeat & 0.771 & 0.727 & 0.756 & 0.775 & 0.747 & \underline{\color[HTML]{0000FF} 0.780} & 0.751 & 0.751 & 0.771 & 0.726 & 0.727 & 0.737 & 0.776 & 0.776 & 0.749 & 0.772 & {\color[HTML]{FF0000} \textbf{0.815}} \\
JapaneseVowels & 0.981 & 0.984 & 0.962 & 0.960 & 0.946 & 0.984 & 0.962 & 0.962 & 0.943 & 0.959 & 0.959 & 0.984 & 0.989 & \underline{\color[HTML]{0000FF} 0.991} & 0.975 & 0.988 & {\color[HTML]{FF0000} \textbf{0.995}} \\
PEMS-SF & 0.867 & 0.861 & 0.751 & 0.838 & 0.855 & \underline{\color[HTML]{0000FF} 0.896} & 0.751 & 0.884 & 0.809 & 0.827 & 0.839 & 0.809 & 0.860 & 0.859 & 0.893 & 0.891 & {\color[HTML]{FF0000} \textbf{0.931}} \\
SelfRegulationSCP1 & 0.840 & 0.908 & 0.908 & \underline{\color[HTML]{0000FF} 0.925} & 0.860 & 0.918 & 0.873 & 0.898 & 0.917 & 0.911 & 0.921 & 0.887 & \underline{\color[HTML]{0000FF} 0.925} & 0.921 & 0.907 & {\color[HTML]{FF0000} \textbf{0.934}} & 0.898 \\
SelfRegulationSCP2 & 0.528 & 0.522 & 0.533 & 0.572 & 0.536 & 0.572 & 0.505 & 0.511 & 0.550 & 0.561 & 0.510 & 0.544 & 0.561 & 0.583 & 0.578 & \underline{\color[HTML]{0000FF} 0.603} & {\color[HTML]{FF0000} \textbf{0.639}} \\
SpokenArabicDigits & {\color[HTML]{FF0000} \textbf{1.000}} & {\color[HTML]{FF0000} \textbf{1.000}} & 0.712 & 0.981 & 0.971 & \underline{\color[HTML]{0000FF} 0.990} & 0.814 & {\color[HTML]{FF0000} \textbf{1.000}} & 0.974 & 0.965 & 0.976 & {\color[HTML]{FF0000} \textbf{1.000}} & 0.988 & 0.979 & 0.983 & 0.987 & {\color[HTML]{FF0000} \textbf{1.000}} \\
UWaveGestureLibrary & 0.878 & 0.859 & {\color[HTML]{FF0000} \textbf{0.944}} & 0.851 & 0.828 & 0.853 & 0.821 & 0.803 & 0.823 & 0.825 & 0.838 & 0.853 & 0.866 & 0.853 & 0.858 & 0.867 & \underline{\color[HTML]{0000FF} 0.900} \\ \midrule
Ours 1-to-1-Wins & 9 & 8 & 6 & 9 & 10 & 9 & 10 & 8 & 8 & 9 & 9 & 9 & 9 & 9 & 9 & 8 & - \\
Ours 1-to-1-Draws & 1 & 1 & 0 & 0 & 0 & 0 & 0 & 2 & 0 & 0 & 0 & 1 & 0 & 0 & 0 & 0 & - \\
Ours 1-to-1-Losses & 0 & 1 & 4 & 1 & 0 & 1 & 0 & 0 & 2 & 1 & 1 & 0 & 1 & 1 & 1 & 2 & - \\ \midrule
Average accuracy ($\uparrow$) & 0.718 & 0.709 & 0.725 & 0.717 & 0.700 & 0.736 & 0.679 & 0.704 & 0.709 & 0.700 & 0.706 & 0.707 & 0.732 & 0.732 & 0.725 & {\color[HTML]{0000FF} 0.742} & {\color[HTML]{FF0000} \textbf{0.777}} \\
Total best accuracy ($\uparrow$) & 1 & 1 & {\color[HTML]{0000FF} 3} & 0 & 0 & 0 & 0 & 1 & 0 & 0 & 0 & 1 & 0 & 0 & 0 & 2 & {\color[HTML]{FF0000} \textbf{5}} \\
Average Rank ($\downarrow$) & 8.85 & 10.4 & 9.5 & 8.9 & 13.2 & 5.35 & 12.7 & 11.2 & 10.85 & 13 & 11.55 & 10.75 & 5.9 & 5.9 & 8.1 & {\color[HTML]{0000FF} 4} & {\color[HTML]{FF0000} \textbf{2.85}}\\ \bottomrule
\end{tabular}%
}
\end{table*}
% \vspace{-0.1in}

\section{Experiments}
\subsection{Experimental setup and Baselines}

 We use the UEA benchmark datasets to validate the superiority of the proposed TimeMIL. These datasets have various lengths, dimensions, and training/test splits. Please refer to Appendix \ref{appendix:dataset} for the details of these datasets. We conduct two groups of experiments as follows. 
 
 \noindent\textbf{Group 1 experiments.} Following \cite{liu2023todynet,li2021shapenet}, this group of experiments is conducted on selected 26 equal-length datasets from UEA to compare with the recent strong baseline methods of multivariate time series classification, including: \textbf{TodyNet} \cite{liu2023todynet}, \textbf{OS-CNN} and \textbf{MOS-CNN} \cite{tang2022omniscale}, \textbf{ShapeNet} \cite{li2021shapenet}, \textbf{TapNet} \cite{zhang2020tapnet}, \textbf{WEASEL+MUSE} \cite{schafer2017multivariate}, \textbf{WLSTM-FCN} \cite{karim2019multivariate}. We also include several well-known traditional methods based on distance and the nearest neighbor classifier, including \textbf{ED-1NN}, \textbf{DTW-1NN-I}, and \textbf{DTW-1NN-D}. Please refer to \cite{liu2023todynet} or Appendix \ref{appendix:implement} for the details of these three baselines. 
 
\noindent\textbf{Group 2 experiments.} Following \cite{anonymous2023moderntcn,wu2022timesnet}, this group of experiments is conducted on selected 10 UEA datasets to compare with the recent strong methods in the general time series analysis, including: \textbf{ModernTCN} \cite{anonymous2023moderntcn}, \textbf{PatchTST} \cite{nie2023a}, \textbf{Crossformer} \cite{zhang2023crossformer}, \textbf{Flowformer} \cite{Flowformer}, \textbf{FEDformer} \cite{zhou2022fedformer}, \textbf{Rlinear} and \textbf{RMLP} \cite{li2023revisiting}, \textbf{MTS-Mixer} \cite{li2023mts}, \textbf{LightTS} \cite{zhang2022less}, \textbf{Dlinear} \cite{zeng2023transformers}, \textbf{TimesNet} \cite{wu2022timesnet}, \textbf{MICN} \cite{wang2023micn}, \textbf{SCINet} \cite{liu2022scinet}, \textbf{Rocket} \cite{dempster2020rocket}, \textbf{LSSL} \cite{gu2022efficiently}, and \textbf{LSTNET} \cite{lai2018modeling}.

% \noindent\textbf{Additional experiments.} In Appendix \ref{appendix:add_results}, we add the results of the proposed method and previous methods in the UEA benchmarks (i.e. all 30 datasets.) %(i.e. both classification-specific methods and general time series frameworks, whose codes are publicly available)

%For the sake of fair and convenient comparison, t
\noindent\textbf{Implementation details.} The previous benchmark results of all baseline methods are taken from their papers, and the same training setting is used.
In the implementation of our proposed model, we adopt the model in \cite{ismail2020inceptiontime} as our backbone, where the output dimension $L$ is fixed to 128. Refer to Appendix \ref{appendix:implement} for details.

%, we use AdamW optimizer \cite{} with a fixed learning rate 1e-3. We use a warmup technique for the first 10 epoch training. We use

\noindent\textbf{Evaluation metrics.} Following \cite{liu2023todynet,li2021shapenet}, we evaluate the performance of our proposed method and other methods by computing the accuracy, average accuracy, average rank, and the number of pair-wise Wins/Draws/Losses.

\vspace{-0.1in}
\subsection{Main Experimental Results}
% \vspace{-0.1in}

The proposed method demonstrates superiority over other recently proposed competing methods in both Group 1 and Group 2 experiments 
(refer to Table \ref{tab:result1} and \ref{tab:result2}). Please refer to Appendix \ref{appendix:add_results} for additional results. 

% In Appendix \ref{appendix:add_results}, we add the results of the proposed method and previous methods in the UEA benchmarks (i.e. all 30 datasets.) 

\begin{figure*}[!t]
\centering\includegraphics[width=0.99\textwidth]{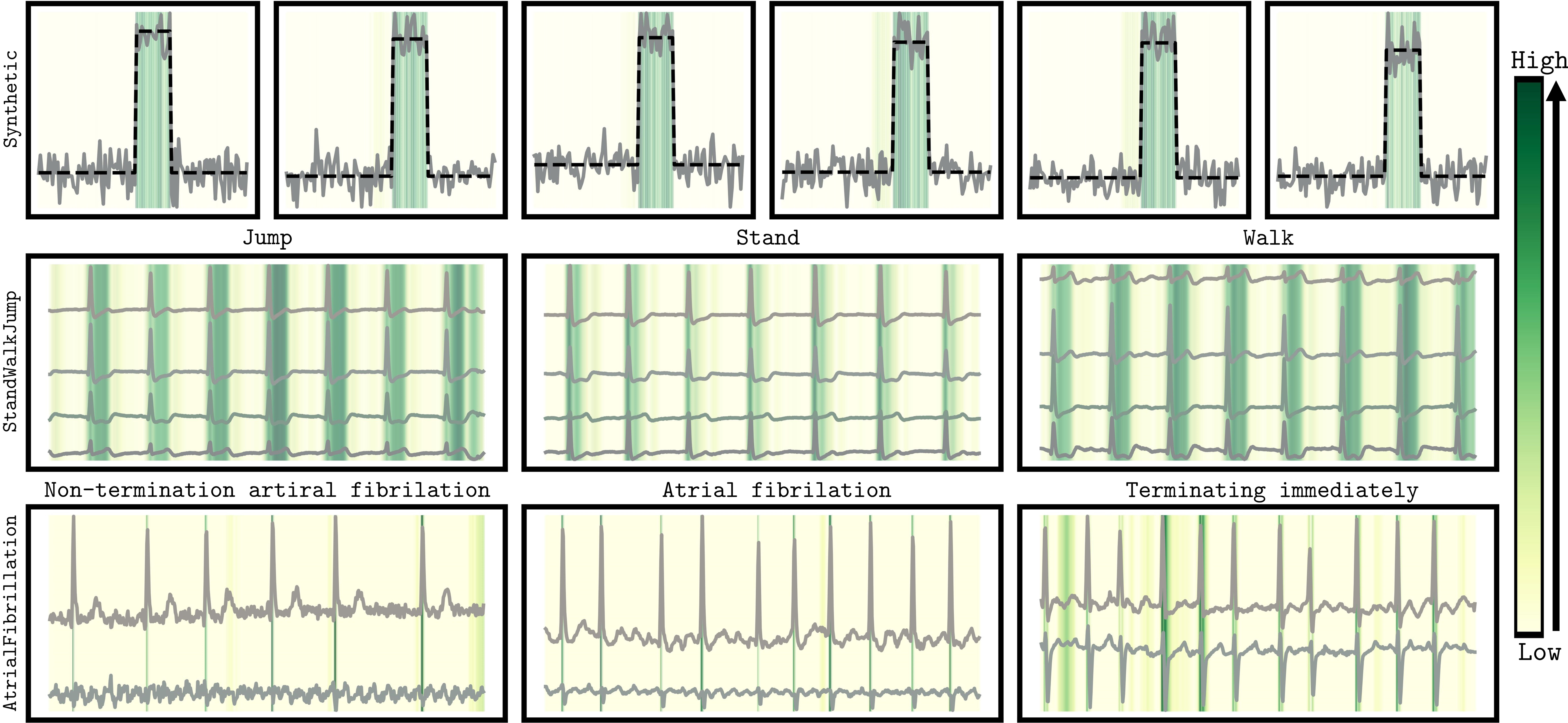}
\vspace{-0.2cm}
\caption{Exemplary attention maps learned in TimeMIL using different datasets (\textbf{rows}) including synthetic dataset, StandWalkJump dataset, and AtrialFibrillation dataset, featuring distinct patterns of interest (\textbf{columns}). TimeMIL accurately localized patterns of interest.}
%\vspace{-0.1in}
\label{fig:attn_map}
\vspace{-0.1cm}
\end{figure*}

\noindent \textbf{Group 1 results.} We obtain a 77.4\% average accuracy on all 26 MTSC datasets, surpassing other methods on 18 datasets and achieving an average rank of 2.327 out of a total of 11 methods. Specifically, 
the proposed method outperforms the second-best methods on each dataset by an average of 4.8\% in accuracy and reduces the performance rank by 1.73. 
% , which achieved 72.6\% and 4.058/11 in average accuracy and rank. 
This performance gain is even more substantial in those challenging datasets.
Specifically, compared to the second-best performing methods, our method improves the average accuracy by 26\%, 20\%, 15\%, 8\%, 8\%, 7\%, and 7\% on {AtrialFibrillation}, {StandWalkJump}, {PEMS-SF}, {FingerMovements}, MotorImagery, FaceDetection, and LSST datasets, respectively.\\
\noindent \textbf{Group 2 results.}
The proposed method also achieved superior performance on the Group 2 datasets compared to other methods.
Specifically, our method achieved a 77.7\% average accuracy, surpassing the other methods in 5 out of 10 datasets and achieving an average rank of 2.85 out of a total of 17 methods. Remarkably, the proposed method outperforms the recent state-of-the-art ModernTCN \cite{anonymous2023moderntcn} by 3.5\% in average accuracy and 1.15 in average rank.

\begin{table}[!t]
\vspace{-0.2cm}
    \centering
     \caption{Comparison of different MIL pooling (\textbf{Left}) and positional encoding (\textbf{Right}) with Group 1 experiments. }
    \begin{minipage}{0.36\columnwidth}\label{tab:pooling}
        \centering
        \resizebox{1\columnwidth}{!}{
        \begin{tabular}{l|c}\toprule
\multicolumn{1}{c}{\textbf{MIL Pooling}} & \multicolumn{1}{c}{\textbf{Accuracy}} \\ \midrule
Mean & 0.715 \\
Max & 0.719 \\
Attention & 0.739 \\
Conjunctive  & 0.746 \\
\textbf{Time-aware (Ours)} & \textbf{0.774} \\ \bottomrule
\end{tabular}%
% \vspace{-0.1in}
}
        % \caption{First Table}
    \end{minipage}%
    \hfill
    \begin{minipage}{0.56\columnwidth}\label{tab:pe}
        \centering
        \resizebox{1\columnwidth}{!}{
        \begin{tabular}{l|cc}\toprule
\multicolumn{1}{c}{\multirow{2}{*}{\textbf{Positional Encoding}}} & \multicolumn{2}{c}{\textbf{Accuracy}} \\
\multicolumn{1}{c}{} & \textbf{ABMIL} & \textbf{TimeMIL} \\ \midrule
None & 0.739 & 0.761 \\
Sinusoidal  & 0.741 & 0.763 \\
\textbf{WPE (Ours)} & 0.745 & \textbf{0.774} \\ \bottomrule
\end{tabular}%
% \vspace{-0.1in}
}
        % \caption{Second Table}
    \end{minipage}
\vspace{-0.13cm}
\end{table}

% \begin{figure*}[!t]
% \centering\includegraphics[width=0.99\textwidth]{figs/attn_all.jpg}
% \vspace{-0.2cm}
% \caption{Exemplary attention maps learned in TimeMIL using different datasets (\textbf{rows}) including synthetic dataset, StandWalkJump dataset, and AtrialFibrillation dataset, featuring distinct patterns of interest (\textbf{columns}). TimeMIL accurately localized patterns of interest.}
% %\vspace{-0.1in}
% \label{fig:attn_map}
% \vspace{-0.1cm}
% \end{figure*}

% \setcounter{figure}{4}
% \begin{figure}[]
%     \centering
%     \includegraphics[width=0.45\textwidth]{figs/decision_boundary.jpg}
%     \vspace{-0.2cm}
%     \caption{Decision boundary learned in fully supervised method (\textbf{Left}) versus TimeMIL (\textbf{Right}) using the synthetic dataset. }
%     \label{fig:decision_boundary}
%     \vspace{-0.2cm}
% \end{figure}

\vspace{0.01cm}
\subsection{Ablation on Model Design Variants}
\noindent\textbf{Effectiveness of TimeMIL pooling.} We compare the performance of the proposed TimeMIL with other commonly used MIL pooling methods, including MeanMIL, MaxMIL, ABMIL \cite{ilse2018attention} and the most recent ConjunctiveMIL \cite{anonymous2024inherently}. We observe that learnable pooling methods (i.e., TimeMIL, ABMIL, and ConjunctiveMIL) show superior performance over non-parametric MIL pooling methods (i.e., MeanMIL and MaxMIL) (Table~\ref{tab:pooling} \textbf{Left}). Notably, the proposed TimeMIL outperforms ABMIL and ConjunctiveMIL by 3\% and 2.8\% in terms of average accuracy, respectively. This supports our initial claim that modeling dependencies between time points is beneficial for MTSC. 

% and the results are shown in the following table. The results demonstrate our method at least has a 3\% gain over other pooling methods, which confirms the substantial superiority of our proposed methods. We conjecture this gain is due to the fact that we model the inherent time step correlation in MIL.
% \begin{table}[]
% \centering
% \label{tab:aba1}
% \resizebox{0.5\textwidth}{!}{%
% \begin{tabular}{l|l}\toprule
% \multicolumn{1}{c}{\textbf{Pooling}} & \multicolumn{1}{c}{\textbf{Accuracy}} \\ \midrule
% Mean & 0.715 \\
% Max & 0.719 \\
% Attention & 0.739 \\
% \textbf{TimeMIL (Ours)} & \textbf{0.774} \\ \bottomrule
% \end{tabular}%
% \begin{tabular}{l|cc}\toprule
% \multicolumn{1}{c}{\multirow{2}{*}{\textbf{Position   Encoding}}} & \multicolumn{2}{c}{\textbf{Accuracy}} \\
% \multicolumn{1}{c}{} & \textbf{Attention} & \textbf{Ours} \\ \midrule
% None & 0.739 & 0.761 \\
% Sinusoidal  & 0.741 & 0.763 \\
% \textbf{WPE (Ours)} & 0.745 & \textbf{0.771} \\ \bottomrule
% \end{tabular}%

% }
% % \vspace{-0.3in}
% \end{table}

\noindent\textbf{Effectiveness of wavelet positional encoding.} We compare the proposed WPE with commonly used Sinusoidal PE \cite{crossattention2}. We observe that adding PE generally improves the performance of both ABMIL and TimeMIL (see Table~\ref{tab:pe} \textbf{Right}), which aligns with our hypothesis that incorporating PE into MIL can better model the ordering within time points and hence lower the classification error. However, we do not observe a significant performance gain ($\sim$ 0.27\%) by adding a sinusoidal PE for both ABMIL and TimeMIL.
On the contrary, the addition of the proposed WPE improves ABMIL and TimeMIL by 0.6\% and 1\%, respectively.
This may be attributed to the fact that it is challenging for predefined sinusoidal PE to capture the changing of frequencies over time within a time series. While the proposed WPE better models these time-frequency changes.

\vspace{-0.1cm}
\subsection{The Effectiveness of Weakly Supervised Learning}
\begin{figure}[t]
    \centering
    \includegraphics[width=0.47\textwidth]{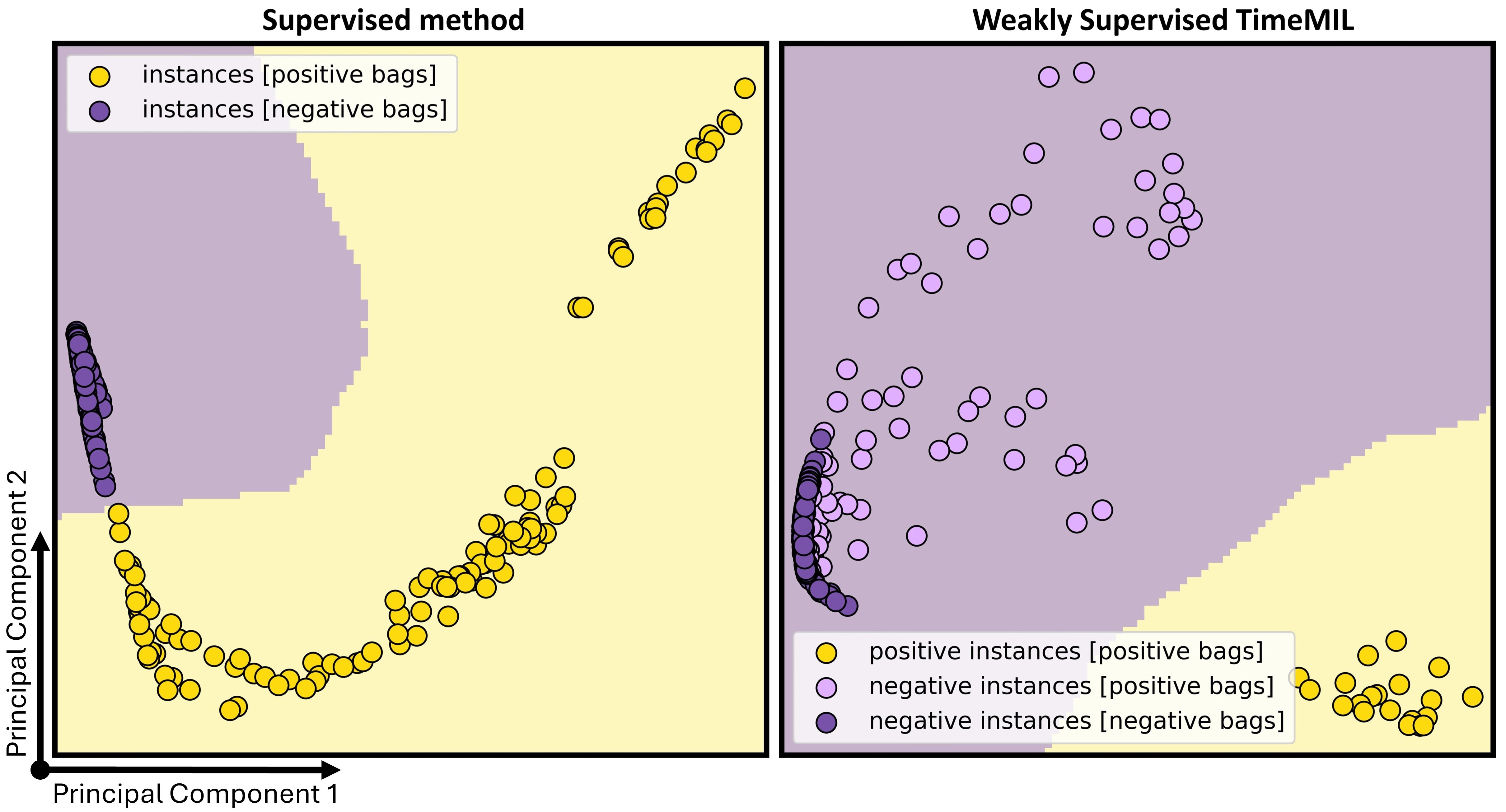}
    \vspace{-0.2cm}
    \caption{Decision boundary learned in fully supervised method (\textbf{Left}) versus TimeMIL (\textbf{Right}) using the synthetic dataset. }
    \label{fig:decision_boundary}
    \vspace{-0.2cm}
\end{figure}

% \vspace{-0.1in}
To validate the effectiveness of the proposed weakly supervised learning scheme, we provide an in-depth comparison between the proposed TimeMIL and traditional fully supervised methods. The results are shown in Fig.~\ref{fig:attn_map} and~\ref{fig:decision_boundary}.

\noindent \textbf{Decision boundary.} We visualize the decision boundary learned in weakly supervised TimeMIL and those learned in the fully supervised method using the synthetic dataset. For the implementation of the supervised method,  we maintain the same feature extractor but replace the TimeMIL pooling with a supervised classifier~\cite{ismail2020inceptiontime}. For visualization of the decision boundary, we projected the time series data onto 2D space by performing PCA on the latent space (i.e., the feature embeddings after applying the feature extractor). More details of the synthetic dataset and visualizing the decision boundary can be found in Appendix~\ref{appendix:Synthetic}.

% \pq{Describe how the decision boundary was generated. How to project the features using PCA.}

We observe that although the fully supervised method could differentiate positive and negative time series, there is not an apparent margin in the decision boundary between the positive and negative time series (see Fig.~\ref{fig:decision_boundary} (\textbf{Left})). This is because the negative instances in positive bags (time series) resemble those negative instances in negative bags. Meanwhile, the positive instances in positive bags are typically less than negative instances. In contrast, the decision boundary of the weakly supervised TimeMIL shows a distinct decision boundary with large margins between the positive and negative instances (Fig.~\ref{fig:decision_boundary} (\textbf{Right})). This unique feature of the proposed TimeMIL provides an instance differentiation between positive and negative time series, offering a precise localization of positive instances.

\noindent \textbf{Inherent interpretability.} 
The instance-level decision-making mechanism makes the proposed TimeMIL inherently interpretable. Leveraging this, we can obtain the importance score.  TimeMIL accurately localized patterns of interest (i.e., positive instances) in the UEA dataset (Fig.~\ref{fig:attn_map}; \textbf{$2^{nd}$ and $3^{rd}$ row}) and synthetic dataset (Fig.~\ref{fig:attn_map}; \textbf{$1^{st}$ row}). This supports our initial hypothesis that time points of interest in positive time series are typically sparse and localized, making the weakly supervised TimeMIL a natural choice for MTSC. 

\section{Conclusion}
% \vspace{-0.1in}
In this work, we introduce TimeMIL, a weakly supervised MIL framework designed for multivariate time series classification. 
The proposed method, from the perspective of weakly supervised learning, offers a better capability to characterize the decision boundary for MTSC than commonly used fully supervised methods.
As a result, TimeMIL demonstrates superiority over 26 methods in 28 datasets and illustrates impressive interpretability by accurately localizing patterns of interest. 
% \vspace{-0.1in}
% In this work, we introduce TimeMIL, a weakly supervised Multiple Instance Learning (MIL) framework designed for multivariate time series classification. The proposed method, from the perspective of weakly supervised learning, offers a better capability to characterize the decision boundary for multivariate time series classification than commonly used fully supervised methods. As a result, TimeMIL demonstrates superiority over 26 methods in 28 datasets and illustrates impressive interpretability by accurately localizing patterns of interest.
% The proposed method can capture the ordering information and temporal dependencies among time points by using a tokenized transformer with a novel wavelet positional encoding. 

% We expect our approach 

% \vspace{-0.1in}
% \section{Limitation}
% \vspace{-0.1in}
%  Currently, our work primarily focuses on modeling the sparsity and locality of time points, and we have not yet extended the interpretability pooling to the channel level. Our work may also be limited by the case that a time point fuses multiple variates that represent potential delayed events or measurements. 
%  We envision that a major difference in expanding to the channel level would involve designing a channel-based positional encoding. This will be a topic of exploration in our future work.
% \vspace{-0.1in}
\section*{Impact Statements}
This paper presents a novel TimeMIL framework to model the temporal correlation and ordering, leveraging a tokenized transformer with a specialized learnable wavelet positional token. The potential societal consequences of TimeMIL can be summarized three-fold: 

\textbf{(i) Theoretical View.} Commencing with a viewpoint in weakly supervised learning, we are the first to formally introduce MIL into time series-related tasks. We rigorously examine feasibility from an information-theoretic perspective. To address a variety of limitations in applying MIL in MTSC, we propose a time-aware MIL pooling to preserve the intrinsic temporal correlation and ordering properties within time series. In summary, we draw theoretical connections between the MIL and MTSC.

\textbf{(ii) Applicability.} Our framework potentially expands the applications of MTSC to financial forecasting, predictive maintenance, anomaly detection, healthcare monitoring, financial forecasting, environmental monitoring, and speech and signal processing. 

% These applications can be explored in future work.

\textbf{(iii) Interpretability.} Compared with the previously existing methods, TimeMIL could provide the interpretable interest of the pattern of the network. By visualizing the points of time series that strongly influence predictions, TimeMIL can assist in ensuring model robustness. This can help identify vulnerabilities and reduce the risk of adversarial attacks. Meanwhile, interpretability contributes to the explainability of MTSC, making it easier for researchers, practitioners, and stakeholders to comprehend/validate how and why a model makes specific predictions.

\textbf{Limitation.} We have not yet extended the MIL to consider the cross-channel information. We envision a major difference in expanding the current framework to the multi-channel version of TimeMIL, which could involve designing cross-channel temporal attention and positional encoding. This will be a topic of exploration in future work.

\section*{Acknowledgement}
This material is based upon the work supported by the National Science Foundation under Grant Number CNS-2204721. It is also supported by our collaborative project with MIT Lincoln Lab under Grant Numbers 2015887 and 7000612889.

% \begin{table}[]
% \centering
% \caption{The NAB score of changing point detection.\textcolor{red}{consider move to appendix.}}
% \label{tab:abnormal_nab}
% \resizebox{0.4\textwidth}{!}{%
% \begin{tabular}{l|ccc}\toprule
% Window size                   & 5              & 10             & 20             \\ \midrule
% Isolation Forest (Individual) & 0.080          & 0.110          & 0.213          \\ 
% Isolation Forest (Mean) & 0.000          & 0.100          & 0.550          \\ \midrule
% \textbf{Our}                  & \textbf{0.550} & \textbf{0.775} & \textbf{0.888} \\ \bottomrule
% \end{tabular}%
% }
% \end{table}


\bibliography{main}
\bibliographystyle{icml2024}

%%%%%%%%%%%%%%%%%%%%%%%%%%%%%%%%%%%%%%%%%%%%%%%%%%%%%%%%%%%%%%%%%%%%%%%%%%%%%%%
%%%%%%%%%%%%%%%%%%%%%%%%%%%%%%%%%%%%%%%%%%%%%%%%%%%%%%%%%%%%%%%%%%%%%%%%%%%%%%%
% \APPENDIX
%%%%%%%%%%%%%%%%%%%%%%%%%%%%%%%%%%%%%%%%%%%%%%%%%%%%%%%%%%%%%%%%%%%%%%%%%%%%%%%
%%%%%%%%%%%%%%%%%%%%%%%%%%%%%%%%%%%%%%%%%%%%%%%%%%%%%%%%%%%%%%%%%%%%%%%%%%%%%%%

\newpage
\appendix
\onecolumn

\section{UEA Datasets Detail}\label{appendix:dataset}
The detail of all 30 datasets is provided in Table \ref{tab:datasets}. It should be noteworthy that the datasets \texttt{JapaneseVowels} and  \texttt{SpokenArabicDigits} used in the Group 2 Experiment originally have varied lengths of sequences. We pre-process it following \cite{wu2022timesnet}, where we pad them to 29 and 93, respectively.

\begin{table*}[hb]
\centering
\caption{Dataset Summary}
\label{tab:datasets}
\resizebox{0.6\linewidth}{!}{%
\begin{tabular}{l|ccccc}
\toprule
\textbf{Dataset} & \textbf{Train Size} & \textbf{Test Size} & \textbf{Dimensions} & \textbf{Length} & \textbf{Classes} \\
\midrule
ArticularyWordRecognition & 275 & 300 & 9 & 144 & 25 \\
AtrialFibrillation & 15 & 15 & 2 & 640 & 3 \\
BasicMotions & 40 & 40 & 6 & 100 & 4 \\
CharacterTrajectories & 1422 & 1436 & 3 & 182 & 20 \\
Cricket & 108 & 72 & 6 & 1197 & 12 \\
DuckDuckGeese & 60 & 40 & 1345 & 270 & 5 \\
EigenWorms & 128 & 131 & 6 & 17984 & 5 \\
Epilepsy & 137 & 138 & 3 & 206 & 4 \\
EthanolConcentration & 261 & 263 & 3 & 1751 & 4 \\
ERing & 30 & 30 & 4 & 65 & 6 \\
FaceDetection & 5890 & 3524 & 144 & 62 & 2 \\
FingerMovements & 316 & 100 & 28 & 50 & 2 \\
HandMovementDirection & 320 & 147 & 10 & 400 & 4 \\
Handwriting & 150 & 850 & 3 & 152 & 26 \\
Heartbeat & 204 & 205 & 61 & 405 & 2 \\
JapaneseVowels & 270 & 370 & 12 & 29 (max) & 9 \\
Libras & 180 & 180 & 2 & 45 & 15 \\
LSST & 2459 & 2466 & 6 & 36 & 14 \\
InsectWingbeat & 30000 & 20000 & 200 & 78 & 10 \\
MotorImagery & 278 & 100 & 64 & 3000 & 2 \\
NATOPS & 180 & 180 & 24 & 51 & 6 \\
PenDigits & 7494 & 3498 & 2 & 8 & 10 \\
PEMS-SF & 267 & 173 & 963 & 144 & 7 \\
Phoneme & 3315 & 3353 & 11 & 217 & 39 \\
RacketSports & 151 & 152 & 6 & 30 & 4 \\
SelfRegulationSCP1 & 268 & 293 & 6 & 896 & 2 \\
SelfRegulationSCP2 & 200 & 180 & 7 & 1152 & 2 \\
SpokenArabicDigits & 6599 & 2199 & 13 & 93 (max)& 10 \\
StandWalkJump & 12 & 15 & 4 & 2500 & 3 \\
UWaveGestureLibrary & 120 & 320 & 3 & 315 & 8 \\
\bottomrule
\end{tabular}}
\end{table*}

% \section{$(\delta_{\varepsilon},\varepsilon)$-continuous symmetric function}
%  Suppose $S:\mathcal{X}\rightarrow\mathbb{R}$. Given $\boldsymbol{X}_i, \boldsymbol{X}_j\in\mathcal{X}$, $\forall d_H(\boldsymbol{X}_i, \boldsymbol{X}_j)<\delta_{\varepsilon}$, we have $ |S(\boldsymbol{X}_i)-S(\boldsymbol{X}_j)|< \varepsilon$.

\section{More detail of Theorem 1}

% \begin{proof}
%    We can apply $\boldsymbol{X}^\prime=\{\sigma(\boldsymbol{x}): \boldsymbol{x} \in \boldsymbol{X}\}$, and $g=S\left(P^{-1}(\cdot)\right)$.
% \end{proof}
We provide an intuition behind this from the perspective of Hausdorff distance, which is defined to measure the difference between two sets from the same feasible domain. Let $(\mathcal{X}, d)$ be a metric space, for each pair of non-empty sets ($\boldsymbol{X}, \boldsymbol{X}') \subset \mathcal{X}$, their Hausdorff distance $d_{\mathrm{H}}$ is computed as 
\begin{align}\label{eq:hausdorff}
d_{\mathrm{H}}(\boldsymbol{X}_i, \boldsymbol{X}_j) \nonumber
:=\max \{&\sup_{\boldsymbol{x}_i^t \in \boldsymbol{X}_i} \inf_{\boldsymbol{x}_j^{t^\prime} \in \boldsymbol{X}_j}  d(\boldsymbol{x}_i^t, \boldsymbol{x}_j^{t^{\prime}}), \\
&\sup_{\boldsymbol{x}_j^t \in \boldsymbol{X}_j} \inf_{\boldsymbol{x}_i^{t^\prime} \in \boldsymbol{X}_i}  d(\boldsymbol{x}_j^t, \boldsymbol{x}_i^{t^{\prime}}) \},
\end{align}
where $\sup$ and $\inf$ denote the supremum operator and the infimum operator, respectively. Eq. \ref{eq:hausdorff} implies that Hausdorff distance measures the furthest distance of traveling from a certain point in a set to its nearest point in the other set under the worst-case scenario. Hence, Hausdorff distance can only measure the distance of time points across different sets and fails to model the time ordering information.

% From an entropy perspective, understanding permutation-variant properties in time series can be quite insightful. As discussed in Section 3.3, we assume that each time point (instance) $\boldsymbol{x}_i^t$ in a time series (bag) is a realization of an R.V $\Theta^t$. The resulting bag R.V. can be represented as $\boldsymbol{X} = \{(\Theta^1|t^1), \cdots, (\Theta^T|t^T)\}$ with the time index $t$, where $t^{j}\neq t^{k},\forall j\neq k$. Here, we use notation $\Theta^j$ for an instance to distinguish a random variable (often as uppercase) and its realization (often as lowercase).  It is fair to construct a general assumption:  
% $p(\Theta^{j}|t^{j})\neq p(\Theta^{j}|t^{k})$, which indicates an instance varies when it is presented in different locations.

% \begin{lemma}
%  For a sequence where the correct knowledge representation necessitates the modeling of permutation information. Under the general assumption, the equality
%  \begin{align}
%  H( \cdots, (\Theta^j|t^j),\cdots) = H(\cdots, (\Theta^j|t^{\Bar{j}}),\cdots),   
%  \end{align} 
%  may \textbf{not} always be satisfied. Here, $\Bar{j}$ is sampled from $[T]$ without replacement.
% \end{lemma}
% \begin{proof}
%     It can be readily proven by contradiction.
% \end{proof}

% In fact, neglecting the permutation information often increases the complexity and predictability of a regular time series, as it often demonstrates some regular or repeating pattern. 

\section{Proof of Proposition 2}\label{appendix:propproof}
\begin{proof}
The existence of equality can be easily proved by assuming the time series is always a constant value, which, regardless of permutation, the entropy is 0.
    The existence of inequality can be proved by contradiction. Suppose a sequence with a length of 2, presenting two tests $\{(\Theta^1|t^1), (\Theta^2|t^2)\}$ for a product. 
    Suppose the probability of passing each test obeys $(\Theta^i)\sim Bernoulli(p= e^{-3+i})$. Suppose the ordering information says the test with index $t^2$ occurs if the product passes the test with index $t^1$.
    we have $p(0,0)=1-e^{-2}, p(0,1)=0, p(1,0)=e^{-2}(1-e^{-1}),  p(1,1)= e^{-1}e^{-2} $.
 
    \begin{align}
        &H(\{(\Theta^1|t^1), (\Theta^2|t^2)\}) \\ \nonumber
        = &-(1-e^{-2})\log(1-e^{-2})-0\log 0 - e^{-2}(1-e^{-1})\log(e^{-2}(1-e^{-1}))- e^{-1}e^{-2}\log( e^{-1}e^{-2}) \\ \nonumber
        =& 0.70~ \text{bit},
    \end{align}

After permuting sequence, $p(0,0)=1-e^{-1}, p(0,1)=0, p(1,0)=e^{-1}(1-e^{-2}),  p(1,1)= e^{-1}e^{-2} $. The entropy of the permuted sequence is presented as, 
\begin{align}
        &H(\{(\Theta^1|t^2), (\Theta^2|t^1)\}) \\ \nonumber
        = &-(1-e^{-1})\log(1-e^{-1})-0\log 0 - e^{-1}(1-e^{-2})\log(e^{-1}(1-e^{-2}))- e^{-1}e^{-2}\log( e^{-1}e^{-2}) \\ \nonumber
        = & 1.16~ \text{bit},
    \end{align}
which are apparently different, and the existence of inequality is proved.
    
    % Accordingly, we can compute the probability of each possibility: $p(0,0)=1-e^{-1}, p(0,1)=0, p(1,0)=e^{-1}(1-e^{-2}),  p(1,1)= e^{-1}e^{-2} $.
    % % The entropy of the original sequence is presented as,
    % \begin{align}
    %     &H(\{(\Theta^1|t^1), (\Theta^2|t^2)\}) \\ \nonumber
    %     = &-(1-e^{-1})\log(1-e^{-1})-0\log 0 - e^{-1}(1-e^{-2})\log(e^{-1}(1-e^{-2}))- e^{-1}e^{-2}\log( e^{-1}e^{-2}) \\ \nonumber
    %     = & 1.16~ \text{bit}.
    % \end{align}
    % If we permute the $t^1$ and $t^2$, we have $p(0,0)=1-e^{-2}, p(0,1)=0, p(1,0)=e^{-2}(1-e^{-1}),  p(1,1)= e^{-1}e^{-2} $.
    %  The entropy of the permuted sequence is presented as,
    % \begin{align}
    %     &H(\{(\Theta^1|t^2), (\Theta^2|t^1)\}) \\ \nonumber
    %     = &-(1-e^{-2})\log(1-e^{-2})-0\log 0 - e^{-2}(1-e^{-1})\log(e^{-2}(1-e^{-1}))- e^{-1}e^{-2}\log( e^{-1}e^{-2}) \\ \nonumber
    %     =& 0.700~ \text{bit},
    % \end{align}
    % which are apparently different, and the existence of inequality is proved.
\end{proof}

\section{Block Entropy}\label{appendix:be}
Block entropy, also known as N-gram entropy, is used to measure the uncertainty of a sequence by \cite{shannon1951prediction}. Suppose a list of overlapping blocks is generated via a sliding window with a size of $n$, where the $j$-th block is $B_j^{(n)}=\left(X_j, \ldots, X_{j+n-1}\right)$. Suppose the set of all appearance of blocks denotes $\left\{b_1^{(n)},\cdots,b_i^{(n),\cdots}\right\}$. For example, in a sequence $AAABBCD$, the set of all possible blocks is $\{AA,AB,BB,BC,CD\}$. 
Then, the block entropy is defined by
\begin{align}
    H_n=-\sum_{i=1}^{L^n} p\left(b_i^{(n)}\right) \log \left(p\left(b_i^{(n)}\right)\right),
\end{align}

% There exist $L^n$ distinct possible blocks of size $n$, which we denote as $\left\{b_i^{(n)}\right\}_{1 \leq i \leq L^n}$. All observed blocks $\left\{B_j^{(n)}\right\}_{j=1, \ldots, N_n}$ in the series $\mathcal{S}$ belong to the set $\left\{b_i^{(n)}\right\}_{1 \leq i \leq L^n}$. The block Shannon entropy is defined by
% $$
% H_n=-\sum_{i=1}^{L^n} p\left(b_i^{(n)}\right) \log \left(p\left(b_i^{(n)}\right)\right), \quad n \geq 1,
% $$
where $p\left(b_i^{(n)}\right)$ denotes the probability of appearance of the block sequence $b_i^{(n)}$. We set $n=2$ in our experiment.

\section{Background of Wavelet Transform}
Mathematically, a wavelet basis $\psi_{a, b}$ is a generated by scaling $a$ and translations $b$ of a single function named \textit{mother wavelet} $\psi\in L^2(\mathbb{R})$, where $L^2(\mathbb{R})$ denotes the Hilbert space of square integrable functions,
\begin{align}\label{eq:wave1}
    \psi_{a, b}(t)=\frac{1}{\sqrt{|a|}} \psi\left(\frac{x-b}{a}\right).
\end{align}
It is noteworthy that the basis $\psi_{a, b}$ is a Hilbert basis, which implies every basis is orthogonal with each other,
\begin{align}\label{eq:wave2}
    \left\langle\psi_{a, b}, \psi_{a^{\prime}, b^{\prime}}\right\rangle \equiv \int_{-\infty}^{\infty}  \psi_{a, b}(t) \psi_{a^{\prime}, b^{\prime}}(t)dt=0.
\end{align}
This also ensures that different wavelet basis are exploring diverse context of the signal.
The \textit{Continuous Wavelet Transform} (CWT) of a 1D signal $f(t)$ is defined by
\begin{align}\label{eq:wave3}
f(a, b)&=\int_{-\infty}^{\infty} f(t) \psi_{a, b}(t) d t= f(t)\circledast \psi_{a, b}(t),
\end{align}
where $\circledast$ denotes the convolutional operation.
We then present the uncertainty principles.
\begin{theorem}\label{th:uncertain}
   \cite{gabor1946theory} Uncertainty principles in the time-frequency version (also known as the \textit{Gabor-Heisenberg limit} ):
\begin{align}
\sigma_t \cdot \sigma_f \geq \frac{1}{4 \pi},
\end{align}
where $\sigma_t$ and $\sigma_f$ denote the measured time and frequency standard deviations, respectively.    
\end{theorem}

\begin{figure*}[htbp]
    \centering
    \includegraphics[width=0.99\textwidth]{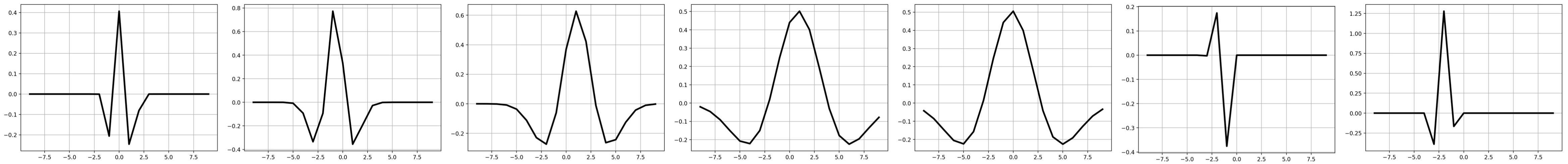}
    \vspace{-0.4cm}
    \caption{The learned wavelet kernel from MTSC tasks.}  %The between-bag distinction measures the between-bag diversity among positive bags.
    \vspace{-0.3cm}
    \label{fig:kernel}
\end{figure*}

We show more diverse learned Wavelet kernel in Fig. \ref{fig:kernel}.

% \section{Main Alogrithm}\label{appendix:alg}

\section{Implementation Detail}\label{appendix:implement}

\subsection{Baselines}
\textbf{ED-1NN}, \textbf{DTW-1NN-I}, and \textbf{DTW-1NN-D} are most popular baselines for MTSC:
% \begin{itemize}
    (i) \textbf{ED-1NN}: It applies the nearest neighbor classifier based on Euclidean distance. 
    (ii) \textbf{DTW-1NN-I}: It applies the nearest neighbor classifier based on dynamic time warping (DTW) that processes each dimension independently.
    and (iii) \textbf{DTW-1NN-D}: It applies the nearest neighbor classifier based on DTW that processes all dimensions simultaneously.
% \end{itemize}

\subsection{More detail about the Architecture}
 We use AdamW optimizer \cite{loshchilov2017decoupled} with a fixed learning rate 1e-3 and a 1e-4 weight decay. We also use Lookahead scheduler \cite{zhang2019lookahead}. Batch sizes are tuned based on the datasets since there are large differences in the dimension and length of each dataset. 
 
As discussed in Section \ref{sec:time_aware}, we use Nyström Self-attention \cite{nys} for accelerating the computation. Specifically, we set the embedding dimension $d_{model}=512$, the number of MHSA heads to $8$. For the Nystrom-based matrix approximation, we set the number of landmark points to 256 and the number of moore-penrose iterations for approximating pseudo-inverse to 6, which is recommended by its original paper. The final classifier consists of two fully connected layers: $\mathbb{R}^L\rightarrow \mathbb{R}^L \rightarrow \mathbb{R}^C$, where $C$ denotes the number of classes. We only feed the class token to the classifier. To facilitate the assumption of TimeMIL that treats MTSC as several \textit{one-vs-rest} (OvR) binary classifications in the context of MIL, we use the binary cross entropy with one-hot encoding for the sequence label. We also adopt window-based random masking augmentation and a warm-up technique, as discussed below. The importance score can be conveniently approximated by using Average-Pooling Based Attention (APBA) proposed by \cite{PDL}.

\subsection{Window-based Random Masking Augmentation}\label{appendix: Masking}
\begin{figure}[htbp]
    \centering
    \includegraphics[width=0.48\textwidth]{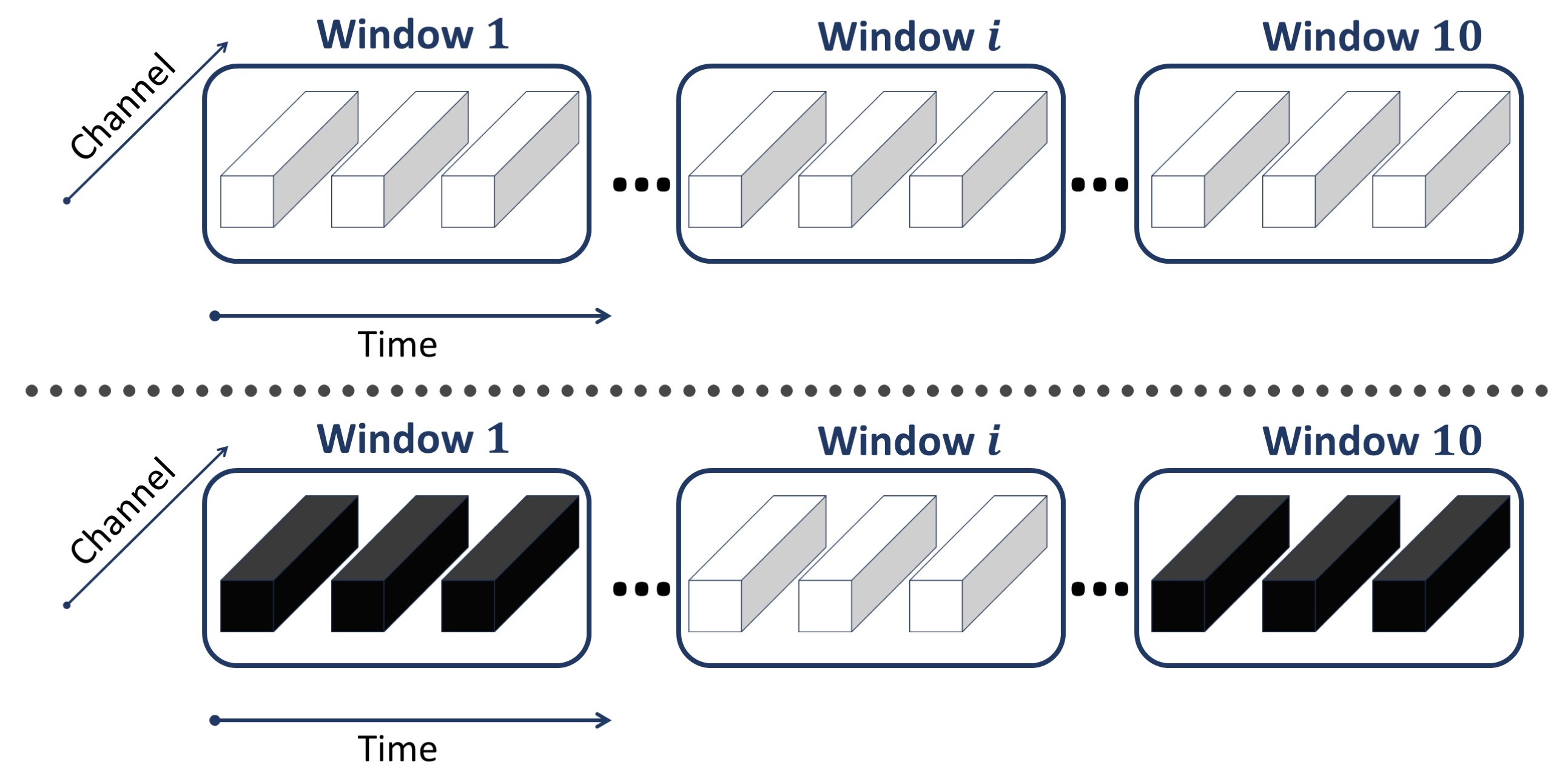}
    \vspace{-0.4cm}
    \caption{The illustration of the window-based random masking. \textbf{Top:} The raw data of the input time sequence. \textbf{Bottom:} An example of the masking, where windows $1$ and $10$ is selected. The masked time points are marked in black. }  %The between-bag distinction measures the between-bag diversity among positive bags.
    \vspace{-0.3cm}
    \label{fig:mask}
\end{figure}

This augmentation is only applied to the raw data in the training phase and aims to lead the model to learn the occlusion-invariant features. Recall the length of the input sequence is $T$. We first generate 10 non-overlapping windows with a size of $T/10$ that can fully cover the entire sequence. Suppose their indices are $[10]=\{1,2,\cdots,10\}$. For each iteration, we sample a set of windows $\mathcal{S}$ with a cardinality $|\mathcal{S}|=10p,p\in (0,1)$ from $[10]$ without replacement. Then, we set the time points covered by the windows in $\mathcal{S}$ to a random noise $\mathcal{N}(0,1)$. The example is shown in Fig. \ref{fig:mask}. We set $p\in\{0,0.5\}$ in our experiment.

% \begin{figure*}[hb]
%     \centering
%     \includegraphics[width=0.99\textwidth]{figs/warmup.jpg}
%     \vspace{-0.4cm}
%     \caption{ }  %The between-bag distinction measures the between-bag diversity among positive bags.
%     \vspace{-0.3cm}
%     \label{fig:warmup}
% \end{figure*}
\subsection{warm-up Training Strategy}\label{appendix: warm-up}
Considering the Transformer-like architecture learns slowly at the beginning, to facilitate the easy gradient flow to the backbone, we apply the following warm-up strategy. Recall the embedding of features is denoted as $\hat{\boldsymbol{x}}_i$, and the class token $\boldsymbol{x}_i^{\operatorname{cls}}$ after applying two transformer blocks. At the first few epochs (empirically set 10), we use $\alpha \mathbb{E}_{t}(\hat{\boldsymbol{x}}_i^t)+(1-\alpha )\boldsymbol{x}_i^{\operatorname{cls}} $ to feed the final classifier, where $\alpha=0.99$. Afterwards, we still use $\boldsymbol{x}_i^{\operatorname{cls}}$ for the final classification.

% \begin{algorithm}
% \caption{Perform dropout on feature set}
% \begin{algorithmic}[1]
% \If{\texttt{args.dropout\_patch} $>$ 0}
%     \State \texttt{select\_window\_idx} $\gets$ \Call{RandomSample}{\texttt{range(10)}, \texttt{int(args.dropout\_patch*10)}}
%     \State \texttt{interval} $\gets$ \texttt{int(len(bag\_feats)/10)}
%     \For{\texttt{idx} in \texttt{select\_window\_idx}}
%         \State \texttt{bag\_feats[:, idx*interval:(idx+1)*interval, :]} $\gets$ \Call{RandomNoise}{}
%     \EndFor
% \EndIf
% \end{algorithmic}
% \end{algorithm}

% \newpage

\section{Additional Experiments}\label{appendix:add_results}
The additional results on all 30 UEA datasets are presented in Table \ref{tab:new_resulst}. 
It is observed that classification-specific methods (TapNet, MOS-CNN, TodyNet, and Ours) often perform better. We also include Mamba \cite{gu2023mamba}, the recent state-of-the-art Selective State Spaces model. We implement the vanilla version of mamba with different numbers of blocks \{1,2,4,8,10,14,22\}, and the results are presented in Table \ref{tab:mamaba}.

% \usepackage{graphicx}
\begin{table}[htbp]
\centering
\caption{Comparison of different methods on 30 UEA datasets. 
(-) indicates the method is not able to obtain results due to memory limitations (we use an A100 40Gb GPU) on the EigenWorms dataset (dimension of 6 and length of 17894) and MotorImagery dataset (dimension of 64 and length of 3000).}
\label{tab:new_resulst}
\resizebox{0.5\textwidth}{!}{%
\begin{tabular}{lccccc}\toprule
\multicolumn{1}{c}{\textbf{Method}} & \textbf{Accuracy} & \textbf{F1} & \textbf{Precision} & \textbf{Recall} & \textbf{AUC-ROC} \\ \midrule
Crossformer(-) & 0.69 & 0.619 & 0.633 & 0.667 & 0.834 \\
PatchTST& 0.702 & 0.664 & 0.694 & 0.686 & 0.854 \\
TimesNet & 0.744 & 0.7 & 0.727 & 0.728 & 0.864 \\
Dlinear & 0.695 & 0.671 & 0.678 & 0.685 & 0.859 \\
FEDformer & 0.701 & 0.664 & 0.697 & 0.682 & 0.867 \\
TapNet & 0.759 & 0.745 & 0.764 & 0.751 & 0.832 \\
MOS-CNN & 0.78 & 0.764 & 0.788 & 0.765 & 0.863 \\
TodyNet & 0.762 & 0.744 & 0.759 & 0.751 & 0.869 \\
Mamba-8 & 0.733 & 0.715 & 0.743 & 0.723 & 0.835 \\
\textbf{Ours} & \textbf{0.791} & \textbf{0.782} & \textbf{0.790} & \textbf{0.782} & \textbf{0.883} \\ \bottomrule
\end{tabular}%
}
\end{table}

% \caption{Classification performance by Mamba with different number of blocks.}
% \label{tab:mamaba}
% \usepackage{graphicx}
\begin{table}[htbp]
\centering
\caption{Classification performance by Mamba with different number of blocks.}
\label{tab:mamaba}
\resizebox{0.5\textwidth}{!}{%
\begin{tabular}{lccccc} \toprule
\multicolumn{1}{c}{\textbf{Method}} & \textbf{Accuracy} & \textbf{F1} & \textbf{Precision} & \textbf{Recall} & \textbf{AUC-ROC} \\ \midrule
Mamba-1 & 0.725 & 0.704 & 0.729 & 0.713 & 0.827 \\
Mamba-2 & 0.726 & 0.705 & 0.729 & 0.715 & 0.83 \\
Mamba-4 & 0.729 & 0.711 & 0.73 & 0.718 & 0.832 \\
Mamba-8 & 0.733 & 0.715 & 0.743 & 0.723 & 0.835 \\
Mamba-10 & 0.727 & 0.705 & 0.727 & 0.714 & 0.833 \\
Mamba-14 & 0.727 & 0.706 & 0.733 & 0.715 & 0.829 \\
Mamba-22 & 0.726 & 0.708 & 0.727 & 0.716 & 0.829 \\
\textbf{Ours} & \textbf{0.791} & \textbf{0.782} & \textbf{0.790} & \textbf{0.782} & \textbf{0.883} \\ \bottomrule
\end{tabular}%
}
\end{table}

\section{Synthetic Dataset}\label{appendix:Synthetic}

\subsection{Dataset Generation}
We simulate a binary dataset similar to noisy pulse signals. Consider the length of a sequence is 120. Again, $\boldsymbol{x}_i^t$ and  ${y}_i^t$ denote the $t$th time point of the sequence $\boldsymbol{X}_i$  and its instance-level label. A negative sequence is generated as,
\begin{align}
    \boldsymbol{x}_i^t\sim \mathcal{N}(0,0.5) \text{ and } {y}_i^t=0.
\end{align}
A positive sequence, which consists of a noisy pulse, is generated as,
\begin{align}
     \boldsymbol{x}_i^t\sim\left\{\begin{array}{l}
 \mathcal{N}(5,0.5) \text{ and } {y}_i^t=1 \text{, for } t\in [a,a+20] \\
\mathcal{N}(0,0.5) \text{ and } {y}_i^t=0, \text{ otherwise}
\end{array}\right.,
\end{align}
where, $a\sim \mathcal{U}(55,65)$, is a random starting point for the pulse signal.

\subsection{Decision Boundary}
We randomly choose a positive sequence (i.e., a bag) and a negative one from the synthetic dataset. After applying feature extractor, both the positive sequence and negative sequence are projected onto a fixed-length (i.e., $L=128$) feature vectors (a positive bag $\boldsymbol{X}_p\in\mathbb{R}^{T\times 128}$ and a negative one $\boldsymbol{X}_n\in\mathbb{R}^{T\times 128}$). In this case, 
we have a total of 240 time points/instances ($T = 240$). Subsequently, we apply PCA for these 240 instances, which reduces their dimensions from 128 to 2 for visualization, meaning we only use the first two principle components. 

% Then, we draw their decision boundaries by using KNN based on their bag label and instance label for the fully supervised method and our weakly supervised method, respectively.

% \section{Additional Way to Obtain Interpretability}
% The interpretability can also be obtained by 
% \newpage
% \bibliography{example_paper}
% \bibliographystyle{icml2024}

\end{document}